\documentclass[oneside,11pt]{article} % For LaTeX2e

\usepackage{amsmath,amsfonts,amssymb,amsthm,commath}

\usepackage{algorithm,algorithmic}

\usepackage[utf8]{inputenc} % allow utf-8 input
\usepackage[T1]{fontenc}    % use 8-bit T1 fonts
\usepackage{microtype}   % better spacing for pdflatex

\usepackage{nicefrac}       % compact symbols for 1/2, etc.

\usepackage{booktabs,enumitem}

\usepackage{graphicx}

\usepackage{url}
\usepackage{xcolor}

\usepackage{hyperref} 

\usepackage{balance} % balance reference

\usepackage{pifont}

\usepackage{dsfont} %%% Specially for the indicator function %\one

\numberwithin{equation}{section}

\newcommand{\R}{\mathbb{R}}

\newcommand{\zeronorm}[1]{\left\lVert #1 \right\rVert_{0}}
\newcommand{\twonorm}[1]{\left\lVert #1 \right\rVert_2}
\newcommand{\onenorm}[1]{\left\lVert #1 \right\rVert_{1}}
\renewcommand{\abs}[1]{\left\lvert #1 \right\rvert}

\newcommand{\nuclearnorm}[1]{\left\lVert #1 \right\rVert_{*}}

\newcommand{\infnorm}[1]{\left\lVert #1 \right\rVert_{\infty}}
\newcommand{\inner}[2]{\langle #1, #2 \rangle}

\renewcommand{\Pr}{\operatorname{Pr}}

\newcommand{\poly}{\mathrm{poly}}

\newcommand{\sign}{\operatorname{sign}}

\newcommand{\X}{\mathcal{X}}
\newcommand{\Y}{\mathcal{Y}}
\newcommand{\D}{\mathcal{D}} %distribution
\newcommand{\DX}{\mathcal{D}_\mathcal{X}} %marginal distribution
\renewcommand{\P}{\mathcal{P}}
\newcommand{\F}{\mathcal{F}} %function class
\newcommand{\M}{\mathcal{M}}
\newcommand{\W}{\mathcal{W}}
\newcommand{\N}{\mathcal{N}}
\newcommand{\EX}{\text{EX}} %oracle
\newcommand{\E}{\mathbb{E}}%Expected value

\newcommand{\one}{\mathds{1}} %mathds
\DeclareMathOperator{\linsum}{GradNorm}
\DeclareMathOperator{\err}{err}
\DeclareMathOperator{\vectorize}{vec}

\DeclareMathOperator{\proj}{proj}
\DeclareMathOperator{\vol}{vol}
\renewcommand{\(}{\left(}
\renewcommand{\)}{\right)}
\theoremstyle{plain}

\ifx\theorem\undefined
\newtheorem{theorem}{Theorem}
\fi

\ifx\lemma\undefined
\newtheorem{lemma}[theorem]{Lemma}
\fi

\ifx\proposition\undefined
\newtheorem{proposition}[theorem]{Proposition}
\fi

\ifx\corollary\undefined
\newtheorem{corollary}[theorem]{Corollary}
\fi

\theoremstyle{definition}

\ifx\remark\undefined
\newtheorem{remark}[theorem]{Remark}
\fi

\ifx\definition\undefined
\newtheorem{definition}[theorem]{Definition}
\fi

\ifx\conjecture\undefined
\newtheorem{conjecture}[theorem]{Conjecture}
\fi

\ifx\fact\undefined
\newtheorem{fact}[theorem]{Fact}
\fi

\ifx\claim\undefined
\newtheorem{claim}[theorem]{Claim}
\fi

\ifx\assumption\undefined
\newtheorem{assumption}{Assumption}
\fi

\usepackage{xspace}

\newcommand{\citet}{\cite}
\title{Attribute-Efficient PAC Learning of Sparse Halfspaces with Constant Malicious Noise Rate}
\author{
Shiwei Zeng\\
Augusta University\\
\texttt{szeng@augusta.edu}
\and
{Jie Shen}\\
Stevens Institute of Technology\\
\texttt{jie.shen@stevens.edu}
}
\begin{document}
\maketitle

\begin{abstract}
Attribute-efficient PAC learning of sparse halfspaces has been a fundamental  problem in machine learning theory. In recent years, machine learning algorithms are faced with prevalent data corruptions or even malicious attacks. It is of central interest to design computationally and attribute-efficient algorithms that are robust to extreme corruptions. In this paper, we consider that there is a constant amount of malicious noise in the data and show that it is possible to PAC learn an underlying $s$-sparse halfspace $w^* \in \mathbb{R}^d$ with $O(s^2\log^5 d)$ samples. Specifically, we follow a recent line of works and assume that the underlying distribution satisfies a concentration condition and a margin condition at the same time. As a complementary result, we provide an information-theoretic sample lower bound under such conditions even for the noiseless case. There is evidence showing that our sample complexity could be nearly optimal. To show the robustness of our algorithm, we provide a new gradient analysis that carefully handles the sparsity admitted constraints in hinge loss minimization program, which could be of independent interest. 
%Our algorirhm follows a simple variant of the existing hinge loss minimization: 1) an attribute-efficient PAC learning algorithm that works under a constant malicious noise rate; 2) a new gradient analysis that carefully handles the sparsity admitted constraints in hinge loss minimization program. 
%Under such conditions, we show that attribute-efficiency can be achieved with simple variants to existing hinge loss minimization programs. Our key algorithmic contribution includes
%As a corollary, our result also implies a simple attribute-efficient algorithm that works under a constant rate of adversarial label noise with lower sample complexity.
%We consider PAC learning of sparse halfspaces that is robust to a constant fraction of malicious noise. PAC learning of halfspaces has been a fundamental problem in machine learning and statistics. In this paper, we revisit this problem and design an attribute-efficient algorithm that learns the underlying $s$-sparse halfspace with $\text{poly}(s,\log d)$ samples and can tolerate up to a constant fraction of malicious noise. Specifically, we follow a recent line of works and assume that the underlying distribution satisfies a certain concentration condition and a margin condition at the same time. To the best of our knowledge, this is the first attribute-efficient algorithm that works under constant malicious noise rate, given that the prior works either is less noise-tolerant or requires $\Omega(d)$ samples.
\end{abstract}

\section{Introduction}

%\Shiwei{Remember to change appendix reference to a blur one.}

%\Shiwei{Remember to include the checklist}

%\Shiwei{remember to check all constant subscripts.}

%{\color{red}
%	The most up-to-date sample complexity is as follows: 
%	%$O(s^2\cdot\log^5 d \cdot\log^2\frac{1}{\delta} +\frac{1}{\eta}\cdot\log\frac{1}{\delta} + \frac{8k}{1-\beta}\log\frac{1}{\delta\epsilon})$
%	\begin{equation*}
%		O\(s^2\cdot\log^5 d \cdot\log^2\frac{1}{\delta} + \frac{1}{\eta_0}\cdot\log\frac{1}{\delta}  +  \frac{8k}{1-\beta}\(\frac{s}{(\epsilon')^2}\cdot \log\frac{2d}{s} +\log\frac{1}{\delta\epsilon}\)     \)
%%		+ \frac{8k}{1-\beta}\log\frac{2}{\delta\epsilon}\)
%	\end{equation*}
%	where $\eta_0\geq\eta$ is the constant upper bound on noise rate.
%}
%Here, the first term is from outlier-removal (Eq.~\ref{eq:sample-soft}); the second part is for that $\abs{S_C}\geq(1-2\eta)\abs{S}$; and the third part is for empirical dense pancake condition from Theorem~\ref{thm:pancake-sample-complexity-restate} with $\epsilon=2\beta'$.
%%\Jie{The $1/\eta$ factor looks incorrect.}

%\Shiwei{Main paper starts here:}

In the modern machine learning and artificial intelligence, designing provably robust algorithms that enjoy desirable data-efficiency has been a pressing problem. One main field that has received  great attention is {\em attribute-efficient learning} which leverages underlying model structures into algorithmic design and analysis such that the total sample complexity of the algorithm depends polynomially on the sparsity parameter $s$ of the underlying model and only polylogarithmically on the ambient data dimension $d$. In this paper, we revisit the fundamental problem of PAC learning of the class of sparse halfspaces, i.e. $\{h_w:x\mapsto\sign(w\cdot x): w\in\R^d,\zeronorm{w}\leq s\}$, and propose an algorithm that enjoys a sample complexity of $O(s^2\log^5 d)$ under extreme noise conditions.
%, where $\epsilon,\delta\in(0,1)$ are accuracy and confidence parameters for PAC learning.

In the past decade, there has been a rich line of works that address PAC learning problems under different noise conditions. Among them, many have achieved attribute-efficiency in algorithmic design for  Massart noise conditions~\cite{awasthi2016learning,zhang2020efficient}, adversarial label noise conditions~\cite{awasthi2016learning,shen2021power}, and malicious noise conditions~\cite{shen2021attribute} under a variety of learning scenarios. However, for challenging noise models such as adversarial lable noise and malicious noise, the best known noise tolerance of algorithms are $\Theta(\epsilon)$, where $\epsilon$ denotes the error parameter, 
%However, for noise models where the adversary is able to corrupt arbitrary samples at its choice, i.e. the adversarial label noise and the malicious noise, the best known noise tolerance is $\Theta(\epsilon)$, where $\epsilon$ denotes the error parameter, 
until the recent works of~\cite{talwar2020error,shen2025efficient} that combined two distributional assumptions of the concentration and large margin. Efficient algorithms tolerant up to a constant amount of adversarial noise~\cite{talwar2020error} and malicious noise~\cite{shen2025efficient} are proposed. Here, we formally define the problem of PAC learning of sparse halfspaces under the malicious noise condition.
%, where the discussion for adversarial label noise is deferred to the appendix.

\begin{definition}[Learning sparse halfspaces with malicious noise]
Let $\X:=\R^d$ be the instance space and $\Y:=\{-1,+1\}$ be the label space.
Let $\EX(\D, w^*, \eta)$ be an adversary oracle with underlying distribution $\D$ over $\X\times\Y$, ground truth $w^*$ with $\zeronorm{w^*}\leq s$, and noise rate $\eta\in(0,\frac12]$ fixed before the learning task begins, such that for all $(x,y)\sim\D, y=\sign(w^*\cdot x)$. Everytime the learner requests a sample from $\EX(\D, w^*, \eta)$, with probability $1-\eta$, the oracle returns a clean sample $(x,y)$ from $\D$;
%the oracle returns an $(x,y)$ where $(x,y)\sim\D$ and $y=\sign(x\cdot w^*)$; 
with probability $\eta$, the oracle returns an arbitrary sample $(x,y)\in\X\times\Y$.  Given any $\epsilon,\delta\in(0,1)$, any $s < d$, the goal of the learner is to output a halfspace $\hat{w}$ using a number $n\leq\poly(s,\log d)$ of samples such that the error rate $\err_{\D}(\hat{w})\leq\epsilon$ with probability $1-\delta$ (over the choice of the samples and all internal random bits of the learning algorithm), where $\err_{\D}(w):=\Pr_{(x,y)\sim\D}(y\neq\sign(w\cdot x))$.
%Given any $\epsilon,\delta\in(0,1)$, $s < d$, and given an adversary oracle $\EX(\D, w^*, \eta)$ with underlying distribution $\D$, ground truth $w^*$ with $\zeronorm{w^*}\leq s$, and noise rate $\eta\in(0,\frac12]$ fixed before the learning task begins, everytime the learner requests a sample from $\EX(\D, w^*, \eta)$, with probability $1-\eta$, the oracle returns an $(x,y)$ where $(x,y)\sim\D$ and $y=\sign(x\cdot w^*)$; with probability $\eta$, the oracle returns an arbitrary sample $(x,y)\in\X\times\Y$. The goal of the learner is to output a halfspace $\hat{w}$ using a number $n=\poly(s,\log d)$ of samples such that the error rate $\err_{\D}(\hat{w})\leq\epsilon$ with probability $1-\delta$, where $\err_{\D}(w):=\Pr_{(x,y)\sim\D}(y\neq\sign(w\cdot x))$.
\end{definition}

In this paper, we propose the first attribute-efficient  algorithm for the above learning problem for any $\eta$ upper bounded by a constant $\eta_0$. Specifically, we follow the algorithmic framework of~\cite{shen2025efficient} and refine the optimization program with sparsity admitted constraints that are extensively used in compressed sensing and regression literature~\cite{tibshirani1996regression,candes2008intro,plan2013robust,plan2016generalized}. That is, we incorporate an $L_1$ norm constraint  as a relaxed condition for the $s$-sparse $w$, and look for an appropriate vector $w$ within the constraint set  $\W:=\{w:\twonorm{w}\leq1,\onenorm{w}\leq\sqrt{s}\}$ that enjoys a small hinge loss on the empirical sample set. We show that the gradient conditions on the optimum $\hat{w}$ ensure its correctness. In more details, we carefully analyze the gradient conditions based on the new constraint set and balance out the influence from both the $L_2$ and $L_1$ constraints. As a result, we show that with the modified constraints, when the underlying halfspace is indeed sparse, our algorithm only requires a sample complexity that depends polynomially in $s,\log d$. We summarize our results in Section~\ref{subsec:main-result}.

%More details can be found in Algorithm~\ref{alg:main}. To mitigate the malicious noise, we also include a soft outlier removal scheme (Algorithm~\ref{alg:soft}) that ensures the weighted variance of the empirical samples are well bounded with respect to all vectors in $\W$.

%Learning under noise conditions is a pressing question in modern machine learning and artificial intelligence.

%On the other hand, data efficiency is crucial in many learning tasts as in many applications we do not have sufficient data to train a good model. Hence, we are taking into consideration the underlying model structure in algorithmic design.

%In this paper, we consider PAC learning of sparse halfspaces under the malicious noise condition. We propose the first attribute-efficient learning algorithm that learns the class of sparse halfspaces that is tolerant to a constant malicious noise rate. 

\subsection{Main result}\label{subsec:main-result}

%\begin{assumption}[Sparsity]
%The underlying halfspace $w^*$ is such that $\zeronorm{w^*}\leq s$.
%% \Jie{With $\ell_0$ constraint, $\ell_1$ becomes redundant.}
%\end{assumption}

Following the works of~\cite{talwar2020error,shen2025efficient}, our result depends on two distributional assumptions.
%We assume that there exists a $\gamma$-margin in the empirical set drawn from the underlying data distribution $\D$ over $\X\times\Y$. Meanwhile, we assume that the marginal distribution $\DX$ is a mixture of $k$ logconcave distributions, such that each one of them satisfies a nice tail bound.

%We consider that the underlying distribution $\D$ is such that the data is $\gamma^*$-margin separable with probability $1$, and the marginal $\DX$ is a mixture of $k$ logconcave distributions. 
%and the labels are given by the underlying halfspace $w^*$.

\begin{assumption}[Large-margin]\label{ass:dataset-margin}%\label{ass:distribution-margin}
Any finite set $S_C$ of clean samples from $\D$ is $\gamma$-margin separable by the target halfspace $w^*$ for some $\gamma>0$. That is, $\forall (x,y)\in S_C$, $yx\cdot w^*\geq\gamma$.
%The data distribution $\D$ over $\X\times\Y$ is $\gamma^*$-margin separable by  $w^*$ for some $\gamma^*>0$. That is, for any $(x,y)\sim\D$, $yx\cdot w^*\geq \gamma^*$ with probability $1$.
\end{assumption}

\begin{assumption}[Mixture of logconcaves]\label{ass:distribution-marginal}
The marginal distribution is $\DX = \frac{1}{k}\sum_{j=1}^{k}\D_j$, where each $\D_j$ is logconcave with mean $\mu_j$ and covariance matrix $\Sigma_j$, satisfying  $\twonorm{\mu_j}\leq r$ for some $r>0$ and $\Sigma_j \preceq \sigma^2 I_d$ for some $\sigma^2=\frac{1}{d}$.
\end{assumption}
%Based on the above two assumptions, we are able to present our main theorem. 
By utilizing the margin and mixture of logconcaves conditions together, we are able to design an attribute-efficient algorithm that learns the underlying $s$-sparse halfspace under $\Omega(1)$ malicious noise.
%Given the two assumptions, we are able to design an algorithm that attribute-efficiently learns $w^*$ with $\Omega(1)$ noise tolerance.

\begin{theorem}[Main result]\label{thm:main}
For any $\epsilon,\delta\in(0,1)$, 
%$\epsilon\in(0,\frac12),\delta\in(0,1)$, 
given that Assumption~\ref{ass:dataset-margin} and \ref{ass:distribution-marginal} hold for some $\gamma\geq { \frac{8(\log\frac{1}{\epsilon} +1)}{\sqrt{d}}}$, $r\leq2\gamma$, $k\leq 64$, and $\eta \leq \eta_0 \leq \frac{1}{2^{28}}$,
by drawing a set of $\Omega\({s^2}\cdot\log^5d\cdot\log^2\frac{1}{\epsilon\delta}\)$ samples from $\EX(\D,w^*,\eta)$,
%$r\in[\frac32\gamma,2\gamma]$, 
Algorithm~\ref{alg:main} runs in polynomial time and returns a halfspace $\hat{w}$  such that $\err_{\D}(\hat{w})\leq\epsilon$ with probability at least $1-\delta$.
%$\Omega\(s^2\cdot \log^5(nd)\log^2\frac{1}{\delta}+\frac{1}{\eta_0}\cdot\log\frac{1}{\delta} + k\cdot\(\frac{s\log d}{\bar{\gamma}^2} +\log\frac{1}{\delta\epsilon}\) \)$ and $\eta \leq \eta_0 \leq \frac{1}{2^{32}}$
%Given a set $S$ of $n\geq \Omega\(s^2\cdot \log^5(nd)\log^2\frac{1}{\delta}+\frac{1}{\eta_0}\cdot\log\frac{1}{\delta} + k\cdot\(\frac{s\log d}{\bar{\gamma}^2} +\log\frac{1}{\delta\epsilon}\) \)$ samples from $\D$ that satisfies Assumption~\ref{ass:distribution-margin} and \ref{ass:distribution-marginal}, and corrupted by the adversary with malicious noise rate $\eta\leq\eta_0\leq\frac{1}{2^{32}}$, there exists an efficient algorithm such that the following holds. For any $\epsilon,\delta\in(0,1)$, $\gamma\geq { \frac{4(\log\frac{1}{\epsilon} +1)}{\sqrt{d}}+2\bar{\gamma} }$, $r\geq2\gamma$, the algorithm returns a $\hat{w}$  such that $\err_{\D}(\hat{w})\leq\epsilon$ with probability at least $1-\delta$.
\end{theorem}

\begin{remark}[Comparison to existing works]
Both two assumptions we made are equivalent to those in the prior works of~\cite{talwar2020error,shen2025efficient}, while their sample complexity is at least $\Omega(d)$. Theorem~\ref{thm:main} suggests that as long as the underlying $w^*$ has a sparse structure, Algorithm~\ref{alg:main} naturally enjoys attribute-efficiency while maintaining its robustness by incorporating sparsity admitted constraints.
%by adding a sparsity admitted constraint, the algorithm naturally enjoys attribute-efficiency while maintaining its robustness.
\end{remark}

As a complementary result, we prove an information-theoretic sample lower bound for learning under the margin and mixture of logconcaves conditions. That is, even under such favorable distributional assumptions, PAC learning is not easier than that under the distribution-free settings.
\begin{lemma}\label{lem:lower-bound}
Learning of $s$-sparse halfspaces under Assumption~\ref{ass:dataset-margin} for any $\gamma\leq\frac12$ and \ref{ass:distribution-marginal} in the noiseless setting has an information-theoretic sample complexity lower bound of $\Omega(s\cdot(\log\frac{1}{\epsilon}+\log\frac{d}{s}))$.
\end{lemma}
\begin{remark}[Optimality]
The prior work of~\cite{dia2017statistical} shows that recovering a $\Theta(s^{-c/4})$-fraction of attribute corruptions of the Gaussian distributions under the Huber contamination model requires at least $\Omega(s^{2-3c})$ samples for any statistical query algorithms running in time $\Omega( d^{\Omega(c\cdot s^c)})$. This implies a quadratic sample lower bound $\Omega(s^2)$ for computationally efficient learning of linear models, i.e. $w\cdot x$, with malicious noise under the statistical query learning setting. Showing information-theoretic lower bound of $\Omega(s^2)$ for computationally efficient algorithms is still an open problem. This evidence shows that our sample complexity could be nearly optimal up to logarithmic factors.
%That is, even for a vanishing constant $c$, the sample complexity is $\Omega(s^2)$.
\end{remark}

%\begin{remark}[Effective]
%
%\end{remark}

\begin{remark}[Constant noise rate]
Our algorithm is tolerant to an $\eta$ malicious noise rate as long as $\eta\leq\eta_0$ for some constant $\eta_0$. This differs from prior works on attribute-efficient learning with noise rate $O(\epsilon)$~\cite{shen2021attribute}, i.e. becoming extremely weak when $\epsilon\rightarrow0$. With carefully optimization over the constants in proofs, the upper bound of $\eta_0$ can be further improved to $2^{-9}/k^{2}$. Yet, this work focus on achieving the attribute-efficiency and we leave these slight improvements to interested readers.
%We believe that the constant upper bound $\eta_0$ can be further improved with a more careful analysis on all constants in proofs. Yet, our effort in this work is in achieving the attribute-efficiency in algorithmic design, and we did not optimize the constants. However, it remains alluring whether one can achieve an $\eta_0$ close to $\frac12$, i.e. the breakdown point of malicious noise tolerance. 
%we focus on achieving attribute-efficiency in our algorithmic design and did not optimize the constant upper bound on the noise rate, i.e. $\eta_0$. It is possible to carefully handle all constants in our analysis and obtain a better bound.
\end{remark}

\begin{remark}[Adversarial label noise]
Our main results immediately imply a computationally and attribute-efficient algorithm that can tolerate up to a constant amount of adversarial label noise. Note that under the adversarial label noise model, the adversary is only allowed to corrupt the labels while all instances remain untouched. Hence, with even a simpler algorithm, 
%i.e. without introducing the soft outlier removal scheme, 
%i.e. hinge loss minimization without sample reweighting Algorithm~\ref{alg:adversarial},
%{i.e. a hinge loss minimization with respect to sparsity admitted constraints, }
we are able to learn under a constant adversarial noise rate and keep the sample complexity polynomial in $s,\log d$. In the main paper, we focus on the more challenging malicious noise and defer such discussions to Appendix~\ref{sec:adv}.
%As this paper mainly deals with the malicious noise, we include it in Appendix~\ref{sec:adv}.
\end{remark}

\begin{remark}[Non-homogeneous halfspaces]
A salient feature of learning under both the mixture of logconcaves and margin conditions is that, the analysis can be adapted to non-homogeneous halfspaces. Though we mainly prove for the homogeneous case, Theorem~15 of~\cite{talwar2020error} shows that the key condition, i.e. the dense pancake condition, remains to hold under Lipschitz linear transformations. This implies that our main results also hold for attribute-efficient learning of non-homogeneous sparse halfspaces under the mixture of logconcaves and margin conditions.
\end{remark}

%With a small variation in our analysis, we design an attribute-efficient algorithm that can tolerate a constant amount of adversarial label noise. We conclude this guarantee as follows.
%\begin{theorem}[Attribute-efficent learning under adversarial noise]\label{thm:main-adv}
%Given a set $S$ of $n\geq\Omega\(\)$ samples from $\D$ that satisfies Assumption~\ref{ass:distribution-margin} and \ref{ass:distribution-marginal}, and corrupted by with adversarial label noise rate $\eta\leq\eta_0$, there exists an efficient algorithm such that the following holds. For any $\epsilon,\delta\in(0,1)$, $\gamma\geq {\color{red} \frac{4(\log\frac{1}{\epsilon} +1)}{\sqrt{d}}+2\bar{\gamma} }$, $r\geq2\gamma$, the algorithm returns a $\hat{w}$  such that $\err_{\D}(\hat{w})\leq\epsilon$ with probability at least $1-\delta$.
%\end{theorem}

\subsection{Technical contributions}

{\bfseries Gradient analysis over sparse admitted constraints.}\ 
The benefit of having both the concentration and margin conditions at the same time is that, when data samples are sufficiently concentrated, the margin will push a high-density region far away from the decision boundary. Both prior works~\cite{talwar2020error,shen2025efficient} rely on this condition and show that any instance that is surrounded by enough amount of good neighbors is not misclassified by an optimum $\hat{w}$, as the  gradients of the instance itself and its neighbors should contribute to a sufficient weight and push the minimization program to move to a  candidate that must be correct on it. Hence, a key technical component here is the gradient  analysis for any instance that lies in a high-density region.
However, this becomes more challenging when both $L_2$ and $L_1$ constraints are implemented. It is known from the Karush–Kuhn–Tucker (KKT) condition that when the program outputs an optimum $\hat{w}$, its gradient (or subgradient) condition is controlled by both the objective function and the list of all constraint functions. Observing that our program is a hinge loss minimization with respect to a convex set, we can implement the KKT condition in our gradient analysis. However, it is in question how to balance out both the influence from the $L_2$ norm constraint $\twonorm{w}\leq1$ and the $L_1$ norm constraint $\onenorm{w}\leq\sqrt{s}$.

Our key observation is that, when any one of the constraints is active, i.e. the solution is on the boundary, there exists some subgradient $g$ of the objective function that lies in the span of the (sub)gradients of the active constraints. The crucial step is to find a vector $w'$ that can be seen as a component of $w^*-\hat{w}$. Conditioned on that the weight from clean samples is prevalent, if the vector $-g$ fails to point in the direction of $w'$ (bringing $\hat{w}$ further towards $w^*$), then it must be because a boundary condition is reached. As a result, we need a $w'$ that is orthogonal to $g$ to establish a set of contradictory conditions and show that $\hat{w}$ is indeed a good enough halfspace. Technically, we will choose $w'=w^*-\hat{w}\inner{w^*}{\kappa}$, where $\kappa$ is some vector related to the boundary conditions, and make sure that $g\cdot w'=0$. Note that we don't need to know the exact vector $w'$, but its existence is enough for us to establish all contradictory conditions. More details can be found in Lemma~\ref{lem:choice-of-w'} and its proof.
%Note that our hinge loss minimization program involves non-smooth functions that requires analysis of subgradients, which is more complicated
%On the other hand, the gradient of any constraint is a vector that is ``perpendicular'' to the norm constraint set boundary.

{\bfseries Function class covering instead of geometric coverings.}\ Under the attribute-efficient learning regime, the previous technique of spacial covering of the target set $\W$ no longer works. This is because no matter we use $L_1$ balls or $L_2$ balls to cover the set $\W$, the sample complexity will blow up. Instead, we revisit the class setting of function space covering~\cite{anthony2002neural} and utilize the existing result for $\N_\infty$-covering the inner product function space. As a result, the sample complexity can be well bounded by the $L_1$ norm bound on $w$ and $L_\infty$ bound on the distance between instances.

{\bfseries Comparison to single isotropic logconcave settings.}\ There is a large line of works for robust learning of halfspaces with or without attribute-efficiency that depend on an assumption that the marginal distribution is a single isotropic logconcave distribution~\cite{balcan2013active,awasthi2017power,zhang2018efficient,shen2021attribute}. Importantly, the isotropic logconcave itself is a very general class of distributions (including the uniforms and Gaussians). However, it still admits a nice property that the probability density stays the same over all directions, i.e. $w\in\R^d,\twonorm{w}\leq1$. This allows bounding the $L_2$ distance between a candidate halfspace $\hat w$ and $w^*$ with $O(\epsilon)$ for $\err(\hat w)\leq\epsilon$. However, this property does not hold any more for the mixture of logconcaves with distribution means deviating from the origin. Fortunately, we utilize the mixture of logconcave assumption together with the margin to show that under specific conditions, $L_2$ distance still implies the disagreement probability, resulting in a hardness example. See Section~\ref{subsec:lower-bound} and Appendix~\ref{sec:lower-bound-app} for more details.

\subsection{Related works}

%In this section, we include more related works to our paper.  
The earliest study of learning with irrelevant attributes could date back to the last century, when~\cite{littlestone1987learn,blum1990learn,blum1991learn} proposed attribute-efficient algorithms for learning concept classes in the online learning settings. Since then, attribute efficiency was studied in different learning problems, such as learning concept classes~\cite{servedio1999compute,klivans2004toward,long2006attribute,helerstein2007pac}, regression~\cite{tibshirani1996regression}, compressed sensing~\cite{donoho2006,candes2008intro}, basis pursuit~\cite{chen1998atomic,tropp2004greed,candes2005decoding}, and variable selection~\cite{fan2001variable,fan2008high,shen2017iteration,shen2017partial}.
%, to name a few.

While the applications of attribute-efficient learning appears to be broad, learning of the class of sparse halfspaces remains fundamental and elusive especially in the presense of data corruptions. The problem has been extensively studied by both the compressed sensing and learning theory communities. Early works focus on the noiseless or benign noise conditions~\cite{boufounos2008one,plan2013one,plan2013robust,zhang2018efficient}, whereas the recent years have witnessed more sophisticated algorithms for extreme noise conditions~\cite{awasthi2016learning,zhang2020efficient,shen2021power,shen2021attribute}.
% with proper distributional assumptions. 
Even without attribute efficiency, it is known that the best possible noise tolerance for distributionally-independent learning under adversarial label noise or malicious noise is $\Theta(\epsilon)$ (\cite{kearns1988learn,kearns1994toward,awasthi2017power}). Under the assumption of concentrated marginals, such as isotropic Gaussian or logconcave types of concentration, the best known noise tolerance was $O(\epsilon)$ (\cite{diakonikolas2020non,shen2021power,shen2021attribute}) until the work of \cite{talwar2020error} introduced the margin condition together with concentration assumptions, and achieved a constant adversarial label noise tolerance. Later on, \cite{shen2025efficient} provided an improved algorithm that achieved a constant noise rate for the malicious noise. This is the line of research that we follow. In addition, with the additional margin condition, the algorithmic design is much simpler comparing to previous designs. Specifically, the main algorithm is a surrogate loss minimization with respect to convex constraints, which is reminiscent of~\cite{plan2013robust,plan2016generalized}. In other words, under proper yet realistic assumptions, we show that simple algorithms can achieve strong noise tolerance and attribute efficiency at the same time.
%and before this work, the best known noise tolerance even under the logconcave distributional assumptions is $O(\epsilon)$. Hence, it is natural for us to ask whether a better noise tolerance is achievable and if positive, under what conditions.

%It is worth noting the close relationship between compressed sensing and learning sparse concept classes, i.e. the field of one-bit compressed sensing~\cite{,,plan2016generalized}
%
%Later on, the problem was studied in the compressed sensing and learning theory community, known as the one-bit compressed sensing and learning sparse concept classes~\cite{plan2013one,plan2013robust,awasthi2016learning} with benign noise
%
%subtle connections between compressed sensing and concept learning was found, known as the one-bit compressed sensing~\cite{plan2013one,plan2013robust}, and hence many established results from sparse signal processing may provide strong evidence in PAC learning of sparse concept classes~\cite{awasthi2016learning}.
%
%%when \cite{littlestone1987learn} proposed the Winnow algorithm for the online learning setting. Following it is a line of research that focuses on the attribute-efficient PAC learning of concept classes in noiseless settings~\cite{littlestone1987learn,helerstein2007pac} 
%
%and addresses the computational and sample efficiency.
%\cite{tibshirani1996regression,candes2005decoding,helerstein2007pac}

\section{Preliminaries}\label{sec:pre}

%{\bfseries Data distributions and noise model.}\ 
%Our analysis focuses mainly on  the assumption that the underlying mixture of logconcaves are separable, such that a natural margin exists in between the classes.

%Our analysis relies crucially on two assumption as follows. The first one is that for distribution $\D$ over $\X\times\Y$, there exists a $\gamma^*$-margin with respect to the underlying halfspace $w^*$.
%\begin{assumption}[$\gamma$-margin distribution]
%	Let $\D$ be a distribution over $\X\times\Y$. There exists $\gamma^*>0$ such that for an underlying halfspace $w^*$ with $\twonorm{w^*}=1$ and $\zeronorm{w^*}\leq s$, it holds that $\Pr_{(x,y)\sim\D}(yx\cdot w^* \geq\gamma^*)=1$.
%\end{assumption}

%On the other hand, we consider that the instances are drawn from a marginial distribution $\DX$ that is a mixture of $k$ logconcave distributions.
%i.e. $\sum_{j=1}^{k}\D_j$, where each $\D_j$ has mean at $\mu_j$ and covariance matrix $\Sigma_j \preceq \frac{1}{d}I_d$.

Given two vectors $u$ and $v$, denote by $u\cdot v$ and $\inner{u}{v}$ their dot product and inner product, respectively. 

{\bfseries Dense pancake.}\ For any given sample $(x,y)$, the dense pancake is defined as a set of samples lying around it within a certain distance. The distance  is measured with respect to some vector $w\in\R^d$. In this paper, we only consider $w$ with sparse properties. That is, we require the dense pancake condition to hold for any $w\in\W$. Note that this is an essential difference between our dense pancake condition and those were defined in the prior works.
\begin{definition}[Dense pancake condition]\label{def:pancake}
%$(\tau,\rho,\epsilon)$-dense pancake.
For $(x,y)\in\X\times\Y$, given any vector $w\in\W$ and a thickness parameter $\tau>0$, the pancake $\P_{{w}}^\tau(x,y)$ is defined as
\begin{equation*}
	\P_{{w}}^\tau(x,y) := \{(x',y')\in\X\times\Y: \abs{y'x'\cdot w - yx\cdot w} \leq\tau \}.
\end{equation*}
We say that the pancake $\P_{{w}}^\tau(x,y)$ is $\rho$-dense with respect to a distribution $\D$ over $\X\times\Y$ if $\Pr_{(x',y')\sim\D}((x',y')\in\P_{{w}}^\tau(x,y)) \geq\rho$.
%\begin{equation*}
%\Pr_{(x',y')\sim\D}((x',y')\in\P_{{w}}^\tau(x,y)) \geq\rho .
%\end{equation*}
Let $\D_1,\D_2$ be two (not necessarily different) distributions over $\X\times\Y$. We say that $(\D_1,\D_2)$ satisfies the $(\tau,\rho,\beta)$-dense pancake condition if for any vector $w\in\W$, it holds that $\Pr_{(x,y)\sim\D_2}(\P_{{w}}^\tau(x,y)\text{ is }\rho\text{-dense w.r.t. }\D_1)\geq 1-\beta$.
%\begin{equation*}
%\Pr_{(x,y)\sim\D_2}(\P_{{w}}^\tau(x,y)\text{ is }\rho\text{-dense w.r.t. }\D_1)\geq 1-\beta .
%\end{equation*}
\end{definition}
{\bfseries Hinge loss.}\ Given any vector $w$, its hinge loss on any $(x,y)$ is given by $\ell_\gamma(w; (x,y)) := \max\{0,1-yx\cdot \frac{w}{\gamma}\}$, 
%\begin{equation}\label{eq:normalized-hinge}
%	\ell_\gamma(w; (x,y)) := \max\left\{0,1-yx\cdot \frac{w}{\gamma}\right\},
%\end{equation}
for a pre-specified parameter $\gamma$. When $\gamma$ is clear from the context, we omit the subscript.
%$\gamma\leq\gamma^*$.
%Let $f(z)=\max(0,1-z)$. The loss function of a halfspace $w$ such that $\twonorm{w}=1$ on a sample set $(x,y)$ is given by $\ell(w;x,y)=f(yx\cdot \frac{w}{\gamma})$. 
For a sample set $S$, the hinge loss of $w$ on $S$ is denoted as $\ell(w;S):=\sum_{i\in S}\ell(w;(x_i,y_i))$. 
%\begin{equation*}
%\ell(w;S):=\sum_{i\in S}\ell(w;(x_i,y_i)) .
%\end{equation*}
If we are further given a weight vector for the sample set $S$, i.e. $q=(q_1,\dots,q_{\abs{S}})$, denote the weighted hinge loss of $w$ by $\ell(w;q\circ S):=\sum_{i\in S}q_i\cdot\ell(w;(x_i,y_i))$.
%\begin{equation*}
%\ell(w;q\circ S):=\sum_{i\in S}q_i\cdot\ell(w;(x_i,y_i)) .
%\end{equation*}

{\bfseries Norms.}\ Formally, we denote by $\onenorm{v}$ the $L_1$ norm of a vector $v$, by $\twonorm{v}$ the $L_2$ norm, and by $\infnorm{v}$ the $L_\infty$ norm. Specially, denote by $\zeronorm{v}:=\abs{\{i:v_i\neq0\}}$ the cardinality of non-zero elements in $v$. For a matrix $A$, we denote by $\onenorm{A}$ the entrywise $L_1$ norm of $A$, i.e. the sum of absolute values of all its entries; and by $\nuclearnorm{A}$ its nuclear norm. For a symmetric $A$, we use $A\succeq0$ to denote that it is positive semidefinite.

Our work relies crucially on the gradient (and subgradient) analysis of the hinge loss minimization program. The following definition of the gradient norm over a linear summation of the instances from a set $S$ plays an important role in restricting the influence of their (sub)gradients.
%As our result lies heavily on the subgradient analysis of the hinge loss, the linear sum norm plays an important role in restricting the influence of the subgradients of any given set of samples. Since the program is subject to an $\ell_1$ norm constraint, we define the linear sum norm in its $\ell_\infty$ form.
\begin{definition}[Gradient norm]\label{def:linsum}
Given a set $S=\{(x_i,y_i)\}$ and weights $q=(q_1,\dots,q_{\abs{S}})$, the gradient norm of a weighted $S$ is defined as
\begin{equation*}
\linsum(q\circ S) := \sup_{a\in[-1,1]^{\abs{S}}} \sup_{w\in \W} \inner{\sum_{i\in S} a_i q_i x_i}{w}.
%\linsum(q\circ S) := \sup_{a\in[-1,1]^{\abs{S}}} \infnorm{\sum_{i\in S} a_i q_i x_i}.
\end{equation*}
\end{definition}
%\Shiwei{This shall be some norm's dual norm, but I can't figure it out immediately. Recall that $\W$ is with both $L_2$ and $L_1$ constraints.}

{\bfseries Other notations.}\ 
%Denote by $q=(q_1,\dots,q_n)$ a weight vector for a set of $n$ samples. 
When $\D, w^*,\eta$ are clear from the context, we write $\EX$ instead of $\EX(\D,w^*,\eta)$.
We use $[\cdot]$ to denote the set of all positive integer that is smaller or equal to the quantity inside. 
%the square brackets.

\section{Algorithm}\label{sec:alg}

The main algorithmic framework follows from~\cite{shen2025efficient}, however, with carefully designed variations for the underlying sparsity structures and the attribute-efficiency requirements. 

There are two major ingradients in Algorithm~\ref{alg:main}.
%In view of the sparsity of the underlying $w^*$, we first filter the set of samples $S'$ drawn from $\EX$ by a simple threshold on their $L_{\infty}$ norms. That is, for all clean samples that are actually from the underlying distribution $\D$, their $L_\infty$ norms are upper bounded by $r+\sigma\cdot(\log{\abs{S'}d}+1)$ with high probability. As a result, we obtain a filtered sample set $S\subseteq S'$.
%Consider $\delta'=c\cdot\delta$ with a small enough $c>0$.
The first ingredient is a soft ourlier removal scheme that assigns weight $q_i\in[0,1]$ to all instance $i\in S$ such that their weighted variance on any sparse direction $w$ is properly upper bounded. This is the key step that restricts the influence of data corruptions from the malicious noise.
The second ingredient is a hinge loss minimization with respect to a set of $L_2$ and $L_1$ norm constraints. Recall that we have assumed that the underlying halfspace $w^*$ is a unit vector in $\R^d$ with $\zeronorm{w}\leq s$ for some $s<d$. Therefore, we consider a set $\W$ of vectors that is a convex hull of the original hypothesis set, i.e. $\{w:\twonorm{w}=1,\zeronorm{w}\leq s\} \subset \W $. This ensures that $w^*$ lies in the feasible set of the minimization program (Algorithm~\ref{alg:main}, Eq.~\eqref{eq:hinge-loss-minimization}).

%Recall that we have assumed that the underlying halfspace $w^*$ is such that $\zeronorm{w^*}\leq s$. Together with the $\ell_2$ norm bound, we have that $\onenorm{w^*}\leq \sqrt{s}$. With this improved $\ell_1$ bound, we are able to run the hinge loss minimization with embedded sparsity (see Eq~\eqref{eq:hinge-loss-minimization} in Algorithm~\ref{alg:main}).
%Note that we defined the hinge loss function $\ell$ with parameter $\gamma$ that is selected to be slightly smaller than $\gamma^*$ (assumed to be known by the algorithm). Together with the fact that $\onenorm{w^*}\leq {\sqrt{s}}$, it is sure that $w^*$ is in the feasible set.
%%it enables the feasibility of hinge loss minimization program.

\begin{algorithm}
	\caption{Main Algorithm}\label{alg:main}
	\begin{algorithmic}[1]
		\REQUIRE $\EX(D, w^*, \eta)$, target error rate $\epsilon$, failure probability $\delta$, parameters $s,\gamma$.
		\ENSURE A halfspace $\hat{w}$.
		\STATE Draw {$n={ C\cdot {s^2}\log^5\frac{d}{\delta\epsilon} }$} samples from $\EX(D, w^*, \eta)$ to form a set ${S}=\{(x_i,y_i)\}_{i=1}^{n}$ for some large enough constant $C>0$.
%		\STATE Draw {$n'={ C\cdot {s^2}\log^5\frac{d}{\delta\epsilon} }$} samples from $\EX(D, w^*, \eta)$ to form a set ${S'}=\{(x_i,y_i)\}_{i=1}^{n'}$.
%		\STATE Remove all samples $(x,y)$ in $S'$ with $\infnorm{x}\geq r+\sigma\cdot(\log\frac{n'd}{\delta'}+1)$ to form a set $S$.\label{step:filter-infty}
		\STATE Apply Algorithm~\ref{alg:soft} on $S$ and let $q$ be the returned weight vector. 
		\STATE Solve the following weighted hinge loss minimization program:
		\begin{equation}\label{eq:hinge-loss-minimization}
			\hat{w} \leftarrow \underset{\twonorm{w}\leq1, \onenorm{w}\leq {\sqrt{s}}}{\arg\min} \ell_\gamma(w;q \circ S).
		\end{equation}
%		\Shiwei{How to guarantee that $\twonorm{w}$ is large enough to keep the pancake condition? First, the margin must exist for the sake of hinge loss and soft margin. We might need to compare the length of that margin to the thickness of pancake.}
%		where the loss function $\ell$ uses parameter  $\gamma$.
%		\Shiwei{$\kappa = \sqrt{s}$, is currently remained for more general bound for $\ell_1$ norm.}
		\STATE Return $\hat{w}$. %$ \leftarrow \hat{v}/\onenorm{\hat{v}}$.
	\end{algorithmic}
\end{algorithm}

It will be convenient to think of $S$ as a union of a set $S_C$ of clean samples that are indeed drawn from the underlying distribution $\D$ and a set of $S_D$ of samples that are maliciously inserted by the adversary, denoted as $S = S_C \cup S_D$.

\subsection{Soft outlier-removal}

%To tackle the malicious noise, we include a soft outlier removal scheme that is heavily adopted in many machine learning problems~\cite{bibid}. 
\vspace{-0.1in}
The robustness of our algorithm to the malicious noise comes mainly from the soft outlier removal scheme, presented in Algorithm~\ref{alg:soft}. Intuitively, it assigns high weights to the clean samples in $S_C$ and low weights to the corrupted samples in $S_D$, especially to those lying far away from the main group. 
%The main ingredient that tackles the feature corruptions in malicious noise model is a soft outlier removal scheme, 
In more details, let $q=(q_1,\dots,q_n)$ be a weight vector with each $q_i$ controls the weight of sample $i\in S$, where $n=\abs{S}$. The goal of Algorithm~\ref{alg:soft} is to return a vector $q$ such that for all $w\in\W$, $\frac{1}{n}\sum_{i=1}^{n}q_i(w\cdot x_i)^2 \leq \bar{{\sigma}}^2$. This is ensured by Lemma~\ref{lem:variance-bound-restate}. By controling the weighted variance, the algorithm effectively restricts the influence from the malicious samples, as the adversary can not insert an instance with very large gradient to twist the optimization program or to force it to optimize towards a wrong direction. If there exist some of such malicious instances, they will be automatically down-weighted by the soft outlier removal scheme and assigned a smaller $q_i$, as a consequence of violating the weighted variance constraint.
%the soft outlier-removal would have reduced its associated weight $q_i$ to a smaller value.

%We then consider the recovered sample set $S$ from the soft outlier-removal scheme (Algorithm~\ref{alg:soft}). This is allowed by the returned weight vector $q$. The guarantee of Algorithm~\ref{alg:soft} ensures that $\forall w: \onenorm{w}\leq {\sqrt{s}}, \twonorm{w}\leq1$, $\frac{1}{n}\sum_{i=1}^{n}q_i(w\cdot x_i)^2 \leq \bar{{\sigma}}^2$ (Lemma~\ref{lem:variance-bound}). This allows us to restrict the influences from the malicious samples, as the adversary can not insert an instance with very large gradient to twist the optimization program or force it to optimize towards a wrong direction. If there were some such malicious instances, the soft outlier-removal would have reduced its associated weight $q_i$ to a smaller value.

The main challenge in implementing such a soft outlier removal scheme is the $L_1$ constraint in set $\W$. It is known to be NP-hard to find the largest variance of $x_i\in S$ with respect to a $w\in\W$, even without the weight vector $q$. Hence, we relax this problem to a semidefinite program by constructing a set $\M:=\{H:H\succeq0,\onenorm{H}\leq s,\nuclearnorm{H}\leq1\}$, and instead look for a largest value of $\frac{1}{n}\sum_{i=1}^{n}x_i^{\top}Hx$ for $H\in\M$. This is presented in Algorithm~\ref{alg:soft} with weight vector $q$. This kind of relaxation was used in prior attribute-efficiency works (e.g. \cite{shen2021attribute} for localized logconcave denoising), and it is known to be computationally and attribute-efficient at the same time. In addition, by running this relaxed program, the output $q$ is a sufficiently good solution to our original problem. That is, $\frac{1}{n}\sum_{i=1}^{n}q_i(w\cdot x_i)^2$ is well bounded for any $w\in\W$.
Our contribution lies in the analysis of the program for a mixture of logconcaves instead of a single localized isotropic logconcave. We reuse part of the proof from~\cite{shen2021attribute} but have significantly adapted it to the new conditions (see appendix).
%, which is included in the appendix.
% is to efficiently remove outliers while maintaining an attribute-efficient sample complexity. However, the constraints in Eq.\eqref{eq:hinge-loss-minimization} is computationally challenging to solve. To remedy this, we introduce a semi-definite program as a relaxation of the original condition and achieve attribute and computational efficiency at the same time.

%The soft outlier-removal scheme, which is popular in the realm of robust statistics and is heavily adopted in machine learning algorithms. The main goal here is to identify the malicious instances and to down-weight them so as to reduce their influence to the hinge loss minimization program. The key technique used here is to introduce a semi-definite program as a relaxation of the optimization of the instance variance over an $\ell_1$ bound, which reserves sample efficiency.

%\Shiwei{more discussion about SDP relaxation.}

\begin{algorithm}
	\caption{Soft Outlier Removal}\label{alg:soft}
	\begin{algorithmic}[1]
		\REQUIRE Sample set $S=\{(x_i,y_i)\}_{i=1}^{n}$, upper bound on empirical noise rate $\xi = 2\eta_0 \geq \frac{\abs{S_D}}{\abs{S}}$, variance parameter $\bar{\sigma} \leftarrow 2\cdot\sqrt{\(\frac{1}{d}+r^2\)}$.
		\ENSURE A weight vector $q=(q_1,\dots,q_n)$.
		\STATE Find a vector $q$ feasible to the following semidefinite program:
		\[
		\begin{cases}
			0\leq q_i\leq1,\ \forall i\in[n], \\
			\sum_{i=1}^{n} q_i \geq (1-\xi)n , \\
			\sup_{H\succeq0,\onenorm{H}\leq s,\nuclearnorm{H}\leq1} \frac{1}{n}\sum_{i=1}^{n} q_i x_i^{\top}Hx_i \leq \bar{\sigma}^2 .
		\end{cases}
		\]
		
		%\Jie{Last constraint is inefficient to evaluate}
		%\Shiwei{I see. I can translate it into SDP after figuring out the quantity for $\sigma$}
		%\Shiwei{pending sufficient condition for $\onenorm{w}$ by $\onenorm{H}$. pending quantity for $\sigma$}
		\STATE Return $q$.
	\end{algorithmic}
\end{algorithm}

%As a result, since we bound the $\ell_1$ norm of the matrix $H$ by a factor of $s$, we are able to solve Algorithm~\ref{alg:soft} using $O(s^2\log^5d\log^2\frac{1}{\delta})$ samples and output an effective $q$.

\subsection{Attribute-efficient hinge loss minimization}

The last yet the most important ingredient, which is also the main source of our algorithmic robustness towards high noise rate, is the hinge loss minimization over the reweighted samples, i.e. $q\circ S$, with respect to sparse set $\W$. 
Note that with the $L_\infty$ norm filter and soft outlier removal scheme, we are able to identify and remove those samples whose attributes are abnormal. However, there could be a large amount of label noise that remains unattended. Hence, it is essential to  utilize the natural robustness of the hinge loss minimization program to handle these label noise. 
%Note that with the $L_\infty$ norm filter and soft outlier removal scheme, we are able to identify and remove most of the samples whose attributes are abnormal. However, there could still be more subtle attribute corruptions. Moreover, the adversary may insert a large amount of samples whose attributes look normal yet with wrong labels. Hence, it is essential to utilize the natural robustness of the hinge loss minimization program to handle the more subtle noises. 
%alone do not suffice for robust learning under a large malicious noise rate, as they only utilize the feature information for filtering. 

Based upon the soft outlier removal scheme, we  attain a weight vector $q$ with respect to which the reweighted variance $\frac{1}{\abs{S}}\sum_{i\in S}q_i(x_i\cdot w)^2$ is well bounded for any  $w\in\W$. This is sufficient for us to show that within any thick and dense enough pancake, the weights for the clean samples are large enough such that the gradients of the malicious samples may not confuse the minimization program too much. As it will be revealed from our gradient analysis, the gradient norm $\linsum(S_D)$ serves as an upper bound on the sum of  gradients over any set $S_D$ projecting onto any $w\in\W$. 
Hence, with guarantees from Algorithm~\ref{alg:soft}, the hinge loss minimization program at Eq.~\eqref{eq:hinge-loss-minimization} ensures that all label noise is also treated. As a result, we can show that a sufficient amount of samples from the underlying distribution $\D$ will not be misclassified by any returned $\hat{w}$. In other words, the error rate of $\hat{w}$ is  bounded.
%Hence, with guarantees from Algorithm~\ref{alg:soft}, the hinge loss minimization program at Eq.~\eqref{eq:hinge-loss-minimization} ensures that all remaining noises are well treated. As a result, we can show that an enough amount of samples from the underlying distribution $\D$ are not misclassified by $\hat{w}$. In other words, the error rate of $\hat{w}$ is  bounded.
%Hence, if the reweighted $\linsum(q\circ S_D)$ is well bounded, the malicious samples are well attended.
% for any vector $w$ with bounded $L_1$ norm
%(Lemma~\ref{lem:subgradient-malicious})
More technically, we  consider a KKT condition on the optimum $\hat{w}$ of the hinge loss minimization program at Eq.~\eqref{eq:hinge-loss-minimization}, such that any $(x,y)$ that has a dense pancake with respect to $\hat{w}$ would not be misclassified by $\hat{w}$. The main challenge lies in the analysis of the KKT condition with both $L_2$ and $L_1$ constraints. As a result, we utilize the underlying unknown parameters $\lambda_1$ and $\lambda_2$ to balance the influence of both constraints and show that the gradient from the hinge loss function would still point to the right direction. This helps us  establish a sufficient condition for showing the correctness of any $(x,y)$ with a dense pancake.

\vspace*{-0.1in}
\section{Analysis}
We present the gradient analysis in Section~\ref{subsec:subgradient}. Statistical results are in Section~\ref{subsec:statistical}. The proof idea of the lower bound is in Section~\ref{subsec:lower-bound}. We provide a proof sketch of main theorem in Section~\ref{subsec:proof-sketch}.

%We first present the most significant theoretical result in Section~\ref{subsec:subgradient}, which is the gradient analysis for the constrained hinge loss minimization program. 
%We then present our analysis for the density condition and the noise rate condition in Section~\ref{subsec:density}, and show how they together establish a robustness condition for our algorithm. In Section~\ref{subsec:deterministic} and \ref{subsec:statistical}, we summarize all deterministic and statistical results for our algorithm. 
%We provide a proof sketch in Section~\ref{subsec:proof-sketch}.

%We first gather all necessary deterministic conditions for our analysis in Section~\ref{subsec:deterministic}. Then, we present our new gradient analysis in Section~\ref{subsec:gradient}. We conclude the sample complexity and statistical results in Section~\ref{subsec:statistical}. In Section~\ref{label}, we prove our main theorem.
%Section~\ref{subsec:deterministic} includes the deterministic results under which the algorithm successfully runs. In Section~\ref{subsec:statistical}, we conclude the statistical results that provide the deterministic conditions with high probability.

\vspace*{-0.1in}
\subsection{Gradient analysis}\label{subsec:subgradient}
%\subsubsection{Analysis on a single $(x,y)$}\label{subsec:single}
The prior works on establishing algorithmic robustness through surrogate loss minimization for distributions with both concentration and margin conditions crucially rely on an analysis for some special $(x,y)$ that is stated as follows. Given the concentration property of the underlying distribution, there must be a large portion of instances that are surrounded by each other. Then, for any $(x,y)$ that lies in a high-density region and with its neighbors having the same label, the KKT condition for any optimum $\hat{w}$ ensures that such an $(x,y)$ is not misclassified by $\hat{w}$. Hence, the following analysis focuses on such a special $(x,y)$ and especially the subgradient conditions around it.

For any given $(x,y)\in\X\times\Y$, we consider the sample set $S$ as the union of three distinct sets with respect to it, i.e. $S=S_P \cup S_{\bar{P}} \cup S_D,$
%\begin{equation}\label{eq:S-union}
%	S=S_P \cup S_{\bar{P}} \cup S_D,
%\end{equation}
where $S_P$ is the set of clean samples (those in $S_C$) that lie in the pancake of $(x,y)$, i.e. $S_P = S_C \cap \P_{\hat{w}}^{\tau}(x,y)$, $S_{\bar{P}}$ includes the clean samples lying outside the pancake, i.e. $S_{\bar{P}} = S_C \backslash S_P$, and $S_D$ again denotes the set of malicious samples.

\begin{theorem}[Correctness of $\hat{w}$ on good $(x,y)$]\label{thm:local-correctness}
Given that Assumption~\ref{ass:dataset-margin} hold. For any given $(x,y)$, if its pancake $\P_{\hat{w}}^{\tau}(x,y)$ with $\tau\leq {\gamma}/{2}$ is $\rho$-dense with respect to $S_C$ for some $\rho>4\eta_0$, and it holds that %$\sum_{i\in S_C\cap \P_{\hat{w}}^{\tau}(x,y)} q_i > \frac{4}{\gamma}\cdot\linsum(q\circ S_D)$, 
\begin{equation}\label{eq:density-noise-rate}
\sum_{i\in S_C\cap \P_{\hat{w}}^{\tau}(x,y)} q_i > \frac{4}{\gamma}\cdot\linsum(q\circ S_D),
\end{equation}
then $(x,y)$ is not misclassified by $\hat{w}$.
\end{theorem}

The proof utilizes an optimality condition on the output $\hat{w}$ that involves the analysis on the subgradients of both hinge loss function and constraint functions. Recall that the weighted hinge loss is defined as $\ell({\hat{w};q\circ S}) = \sum_{i\in S} q_i \max\{0,1-y_ix_i\cdot\frac{\hat{w}}{\gamma}\}$ (see Section~\ref{sec:pre}). We decompose this loss into three parts, i.e. 
\begin{equation*}
\ell({\hat{w};q\circ S}) = \ell({\hat{w};q\circ S_P}) + \ell({\hat{w};q\circ S_{\bar{P}}}) +\ell({\hat{w};q\circ S_D}) ,
\end{equation*}
where $q\circ S'$ for any subset $S'\subseteq S$ denotes a weighted sample set according to the weight vector $q_{S'} := \{q_i, i\in S'\}$ while ignoring the unrelated weights. 

We prove the theorem by contradicting the optimality of $\hat{w}$ (Algorithm~\ref{alg:main}, Eq.~\eqref{eq:hinge-loss-minimization}) and that a given $(x,y)$ described in Theorem~\ref{thm:local-correctness} is misclassified by $\hat{w}$. While outputing $\hat{w}$, if $\hat{w}$ is the interior of the constraint set $\W:=\{w:\twonorm{w}\leq1,\onenorm{w}\leq {\sqrt{s}}\}$, the analysis follows directly from the prior works.
The analysis for $\hat{w}$ on the boundary is more complicated. Denote by $\hat{z} \in \partial_w\onenorm{w}\mid_{w=\hat{w}}$ a subgradient of $L_1$ constraint function at $\hat{w}$, and it is known that $\{\hat{w}\} = \partial_w\twonorm{w}\mid_{w=\hat{w}}$ is the gradient set for the $L_2$ constraint.
Specifically, due to the KKT conditon,  there exists some $g\in \partial_w \ell(w;q\circ S)\mid_{w=\hat{w}}$ and some $\hat{z}$ such that 
%{\color{red}($\exists/\forall$), ans: $\exists$}
\begin{equation*}
g + \lambda_1\cdot\hat{z}  + \lambda_2\cdot\hat{w}  = 0,
%0 \in \partial_w \ell(w;q\circ S)\mid_{w=\hat{w}} + \lambda_1\cdot \partial_w\onenorm{w}\mid_{w=\hat{w}} + \lambda_2\cdot   \partial_w\twonorm{w}\mid_{w=\hat{w}}
\end{equation*}
where $\lambda_1,\lambda_2\geq0$. More precisely, the complementary slackness implies that $\lambda_1>0$ if and only if $\onenorm{\hat{w}}=\sqrt{s}$ and $\lambda_2>0$ if and only if $\twonorm{\hat{w}}=1$. When only one of the constraints is active, the proof is rather intuitive as the span of a subgradient $g$ is clear, i.e. either due to $\hat{z}$ or $\hat{w}$. It is when both of the constraints are in effect a much more challenging case. Hence, we will show the proof by considering that $\lambda_1,\lambda_2>0$, and the proof for either $\lambda_1=0$ or $\lambda_2=0$ follows as special cases.
%\Shiwei{not 100\% sure, appreciate if you can confirm this.}\Shiwei{2025.04.10: Now I 99\% confirmed.}
%When only the $L_2$ norm constraint $\twonorm{\hat{w}}=1$ is in effect, the analysis is rather intuitive and follows from the prior works. 
%When only the $L_1$ norm constraint $\onenorm{\hat{w}}=\sqrt{s}$ is in effect, we provide new analysis for this and the method is rather intuitive. 
%The most complicated case is that when both two constraints are in effect. We will show the proof by considering the third case, as the first two follows from it by setting either $\lambda_1$ or $\lambda_2$ to $0$.
%%%%%%%%%%%%%%Below is old proof follows from that when there is only the L-1 norm constraint.
%For any $\hat{z}\in  \partial_w\onenorm{w}\mid_{w=\hat{w}}$, define $w' = {\color{red}\sqrt{s}}\cdot \frac{w^* - \hat{w}\inner{w^*}{\hat{z}}}{\onenorm{w^* - \hat{w}\inner{w^*}{\hat{z}}}}$. We will show that if $(x,y)$ is misclassified by $\hat{w}$, $\forall g\in \partial_w \ell(w;q\circ S)\mid_{w=\hat{w}}$we have that 
%\begin{equation*}
%g\cdot w' \leq -\frac14\gamma\sum_{i=1}^{n}q_i + \linsum(q\circ S_D),
%\end{equation*}
%while the optimality of $\hat{w}$ confirms that $\exists g$ such that $g\cdot w' =0$. Similarly, for both constraints are in effect, for any $\hat{z}$, we define $w' = {\color{red}\sqrt{s}}\cdot \frac{w^*-\hat{w}\inner{w^*}{\beta}}{\onenorm{w^*-\hat{w}\inner{w^*}{\beta}}}$, where $\beta = \frac{1-\lambda_2}{\kappa\lambda_1}\cdot\hat{z} + \frac{1-\kappa\lambda_1}{\lambda_2}\cdot\hat{w}$, and show similar contradiction.
%%%%%%%%%%%%%%

The  analysis relies crucially on the a vector $w'$ that can be considered as the component of $w^*-\hat{w}$ that is orthogonal to some subgradient $g$. That is, we hope that $g\cdot w'=0$. To find a vector satisfying this condition, we define
\begin{equation}\label{eq:design-of-w'}
w' := w^*-\hat{w}\inner{w^*}{\kappa},
\end{equation}
where $\kappa = \frac{\lambda_1}{\sqrt{s}\lambda_1+\lambda_2} \cdot\hat{z} + \frac{\lambda_2}{\sqrt{s}\lambda_1+\lambda_2} \cdot\hat{w}$. The following lemma guarantees that there exists some $g$ such that $g\cdot w'=0$. The proof is deferred to the appendix.
\begin{lemma}\label{lem:choice-of-w'}
Consider the set of subgradients $\partial_{w}\ell(w; q\circ S)\big|_{w=\hat{w}}$ and some element $g$ in it. Given the optimization program Eq.\eqref{eq:hinge-loss-minimization}  in Algorithm~\ref{alg:main} and $\hat{w}$ returned by it, let $w'$ be the vector defined in Eq.~\eqref{eq:design-of-w'},
%$w' = w^*-\hat{w}\inner{w^*}{\kappa}$.
%$w' = \frac{w^*-\hat{w}\inner{w^*}{\beta}}{\twonorm{w^*-\hat{w}\inner{w^*}{\beta}}}$. 
%By choosing $\kappa = \frac{\lambda_1}{{\sqrt{s}}\lambda_1+\lambda_2} \cdot\hat{z} + \frac{\lambda_2}{{\sqrt{s}}\lambda_1+\lambda_2} \cdot\hat{w}$ 
for some $\lambda_1,\lambda_2\geq0, \lambda_1+\lambda_2\neq0$ and $\hat{z}\in\partial_w\onenorm{w}\big|_{w=\hat{w}}$. Then, there exists $g\in \partial_{w}\ell(w; q\circ S)\big|_{w=\hat{w}}$ such that $g\cdot w'=0$.
\end{lemma}
%See Lemma~\ref{lem:choice-of-w'} for more details. As a result, we have the following guarantee. The proof is deferred to Section~\ref{subsec:subgradient}.
The following lemma shows that, if an $(x,y)$ is misclassified by $\hat{w}$, then all clean samples within its pancake will contribute to a good weight $\sum_{i\in S_P}q_i$, together with those from the malicious samples, we construct the following upper bound on the dot product $g\cdot w'$ for any subgradient $g$.
\begin{lemma}\label{lem:subgrad-prime}
Consider Algorithm~\ref{alg:main}. Suppose Assumption~\ref{ass:dataset-margin} holds. For some $\hat{z}\in \frac{\partial \onenorm{w}}{\partial w}\big|_{w=\hat{w}}$, define  $w' = w^*-\hat{w}\inner{w^*}{\kappa}$, where $\kappa = \frac{\lambda_1}{{\sqrt{s}}\lambda_1+\lambda_2} \cdot\hat{z} + \frac{\lambda_2}{{\sqrt{s}}\lambda_1+\lambda_2} \cdot\hat{w}$. Then, we have the following holds. For any $(x,y)\in\X\times\Y$ and its pancake $\P_{\hat{w}}^{\tau}(x,y)$ with $\tau\leq\gamma/2$, if $(x,y)$ is misclassified by $\hat{w}$, then for any $g\in\partial_{w}\ell(w; q\circ S)\big|_{w=\hat{w}}$, we have
\begin{equation*}
g\cdot w' \leq -\frac12 \sum_{i\in S_P}q_i + {\frac{2}{\gamma} \cdot} \linsum(q\circ S_D). 
\end{equation*}
\end{lemma}
Notice that when the weight $\sum_{i\in S_P}q_i $ is large enough, i.e. the condition in Eq.\eqref{eq:density-noise-rate} is satisfied, it will push the dot product $g\cdot w'$ towards the negative side, such that $g\cdot w'<0$ holds for all subgradients $g\in \partial_w \ell(w;q\circ S)\mid_{w=\hat{w}}$. Interestingly, this means that the program will keep optimizing towards $w^*$, and is contradictory to the condition we establish in Lemma~\ref{lem:choice-of-w'}. Hence, we prove by contradiction that any $(x,y)$ with a large weight $\sum_{i\in S_P}q_i $ is not misclassified by $\hat{w}$. The guarantees are summarized  in Theorem~\ref{thm:local-correctness} and the full proof is deferred to Appendix~\ref{subsec:subgradient-supp}.

\subsection{Sample complexity}\label{subsec:statistical}

The sample complexity of our algorithm mainly comes from three sources: 1) the amount of samples required for the empirical dense pancake condition, i.e. $(S_C, \D)$ satisfies $(\tau,\rho,\epsilon)$-dense pancake condition with respect to any $w\in\W$;
%Assumption~\ref{ass:pancake-deterministic}, 
%which is a criteria for Algorithm~\ref{alg:main}; 
2) the amount of samples that ensures the statistical properties for Algorithm~\ref{alg:soft}. 
%We first present the sample complexity for the empirical pancake density;
%The sample complexity consists of  two parts: 1) the amount of samples required for Assumption~\ref{ass:dataset-margin} and \ref{ass:pancake-deterministic} to hold, which is a criteria for Algorithm~\ref{alg:main}; 2) the amount of samples we use to ensure the statistical property tested in Algorithm~\ref{alg:soft} is satisfiied.
3) the sample complexity that bounds the empirical noise rate by $\xi=2\eta_0$.

\begin{theorem}[Sample complexity for empirical pancake density]\label{thm:pancake-sample-complexity}
Consider a distribution $\D$ over $\X\times\Y$ satisfying Assumption~\ref{ass:distribution-marginal}  and \ref{ass:dataset-margin} with $\gamma>2\tau$, where $\tau=2\sigma\cdot\(\log \frac{1}{\beta}+1\)$. It holds with probability at least $1-\delta'$ over the draw of $\abs{S_C}\geq \frac{4k}{1-\beta}\cdot\({C_0s}\cdot \log^4{d} +\log\frac{1}{\delta'\beta}\)$ samples, $(S_C,\D)$ satisfies $\(2\tau,\frac{1-\beta}{2k},2\beta\)$-dense pancake condition, where $C_0>0$ is an absolute constant.% and $\rho=\frac{1-\beta}{2k}$. 
%Consider a distribution $\D$ over $\X\times\Y$ satisfying Assumption~\ref{ass:dataset-margin} and \ref{ass:distribution-marginal} with $\gamma^*>\tau$, where $\tau=\tau'+\epsilon'$ and  $\tau'=2\sigma\cdot\(\log \frac{1}{\beta}+1\)$. It holds with probability at least $1-\delta'$ over the draw of $\abs{S_C}\geq \frac{2}{\rho}\cdot\(\frac{C_0s}{(\epsilon')^2}\cdot \log\frac{2d}{s} +\log\frac{1}{\delta'\beta'}\)$ samples, $(S_C,\D)$ satisfies $\(\tau,\rho,\beta+\beta'\)$-dense pancake condition, where $\rho=\frac{1-\beta}{2k}$. 
\end{theorem}
%The proof is deferred to Section~\ref{sec:sample-comp-supp}.
It is straightforward to show that any distribution $\D$ satisfying both Assumption~\ref{ass:distribution-marginal}  and \ref{ass:dataset-margin} is $(\tau,\rho,\beta)$-dense with $\tau=2\sigma\cdot(\log \frac{1}{\beta}+1)$ and $\rho=\frac{1-\beta}{2k}$ (see Lemma~\ref{lem:dense-distribution} in Appendix~\ref{sec:sample-comp-supp}). The challenge here is to show that by collecting a $O(\poly(s,\log d))$ amount of samples from $\D$, the empirical set $S_C$ is $(2\tau,\rho,2\beta)$-dense with respect to $\D$ with high probability. 

The prior works of~\cite{talwar2020error} proves it using standard geometric covering. Specifically, they first show that the dense pancake condition holds for a fixed vector $w$ except for a small failure probability. Then, it suffices to take a union bound of the failure rate over a $\tau$-cover of the unit ball in $\R^d$, since an $L_2$ distance of $\tau$ between $w,w'$ only increase the pancake width by $\tau$ while all other density conditions stay unchanged (see Theorem~20 therein). We remark that this analysis does not work under the sparsity conditions. First, using $L_2$ balls of radius $\tau$ to cover $\W$ yields a covering number of at least $\exp(s/\tau^2)$. For small $\tau=\frac{1}{\sqrt{d}}$, this leads to a blow-up in sample complexity of $\poly(d)$. Moreover, to the best of our knowledge, it remains unknown whether the covering number is still $O(s\log d)$ when using $L_1$ balls of radius $\tau$ to cover the set $\W$.
%That is, to show that the dense pancake condition holds for any $w$ residing in the unit ball in $\R^d$, it suffices to cover the unit ball with small balls of $L_2$ radius of $\tau$.
% can be

Hence, instead of geometric covering, we utilize results from the traditional function space covering and
construct a $\tau$-cover over the inner product space of $\inner{w-w'}{\bar{x}}$, where $\bar{x}$ represents the distance from any point to the pancake center $(x,y)$. It is known that for any logconcave component, samples are away from each other with $L_\infty$ distance of at most $\sigma\log d$ with high probability. We also know that for any pairs of $w,w'\in\W$, the $L_1$ distance is at most $\sqrt{s}$. Hence, it suffices to bound the covering number $\N_{\infty}$ by $\exp\big(\tilde{O}\big(\frac{\sqrt{s}^2(\sigma\log d)^2}{\tau^2}\big)\big)$ using Corollary~5 from~\cite{zhang2002cover} based on the norm bounds on both parts for the inner product. Here, $\tilde{O}$ hides logarithmic terms. Notice that $\N_{\infty}$ differs from $L_\infty$ as it gaurantees the covering of functions to be effective for all instances $x$ in the target sample set.
%That is, we incorporates a result from~\cite{zhang2002cover} that upper bounds the $\infty$-norm covering number of the inner product function space induced by samples with bounded $L_\infty$ norm and linear weights with bounded $L_1$ norm. 
See more details in the full proof of Theorem~\ref{thm:pancake-sample-complexity} (Appendix~\ref{sec:sample-comp-supp}) and in Appendix~\ref{sec:useful_lemmas},

The following lemma dedicates to the sample complexity for the semidefinite program in Algorithm~\ref{alg:soft}. 
%Note that the constrained set in the algorithm is a subset of $\{H:\onenorm{H}\leq s\}$. Hence, the upper bound still works for our set of $H$.

\begin{lemma}[Sample complexity for empirical variance]\label{lem:soft-sample-complexity}
%There exists an absolute value $C_1>0$ such that the following holds 
For any distribution $\DX$ satisfying Assumption~\ref{ass:distribution-marginal}, let $S_C=\{x_1,\dots,x_{\abs{S_C}}\}$ be a set of i.i.d. unlabeled instances drawn from $\DX$. Denote $G(H) := \frac{1}{\abs{S_C}}\sum_{i\in S_C}x_i^{\top}Hx_i - \E_{x\sim\DX}[x^{\top}Hx]$. Then with probability $1-\delta'$, $\sup_{H\in\M}\abs{G(H)} \leq 2\(\frac{1}{d}+r^2\)$
%\begin{equation*}
%\sup_{H\in\M}\abs{G(H)} \leq 2\(\frac{1}{d}+r^2\)
%%\sup_{H:\onenorm{H}\leq s}\abs{G(H)} \leq 2\(\frac{1}{d}+r^2\)
%\end{equation*}
given that $\abs{S_C}\geq {\Theta\(s^2\cdot \log^5 d \log^2\frac{1}{\delta'} \)}$. 
\end{lemma}
Together with the following lemma, we obtain the variance parameter in Algorithm~\ref{alg:soft}. That is, if $\forall H\in\M$, $\frac{1}{\abs{S_C}}\sum_{i\in S_C}x_i^{\top}Hx_i \leq 4\(\frac{1}{d}+r^2\)$ with high probability, it suffices to set $\bar{\sigma}=2\sqrt{\frac{1}{d}+r^2}$ for the program to identify the clean samples and downweight all malicious samples.
%Due to the following lemma, and Lemma~\ref{lem:soft-sample-complexity} in Section~\ref{subsec:statistical}, we note that it suffices to set $\bar{\sigma}\leftarrow 2\sqrt{(1/d+r^2)}$ for the algorithm to idenfity a good $q$ given a reasonable amount of samples.
\begin{lemma}\label{lem:clean-variance}
Given distribution $\DX$ satisfying Assumption~\ref{ass:distribution-marginal}, it holds that $\sup_{H\in\M }\E_{X\sim\DX}(x^{\top}Hx) \leq 2\(\frac{1}{d}+r^2\)$.
%Given distribution $\DX$, it holds that $\sup_{H\in\M}\E_{X\sim\DX}(x^{\top}Hx) \leq 2\(\frac{1}{d}+r^2\)$.
\end{lemma}

The following lemma bounds the empirical noise rate $\frac{\abs{S_D}}{\abs{S}}$, which facilitates our analysis.% for Algorithm~\ref{alg:soft}. 
%It is also important to upper bound the empirical noise rate, i.e. the requirement for Algorithm~\ref{alg:soft}. 
%That is, $\frac{\abs{S_D}}{\abs{S}}\leq2\eta_0$.
%Recall that $\xi=2\eta_0\geq\frac{\abs{S_D}}{\abs{S}}$.
\begin{lemma}[Sample complexity for empirical noise rate]\label{lem:clean-sample-complexity}
	If we draw a sample $S$ from $\EX$ with size $\abs{S}\geq\frac{3}{\eta_0}\cdot\log\frac{1}{\delta'}$, then it is guaranteed with probability $1-\delta'$ that $\abs{S_C} \geq (1-2\eta_0)\abs{S}$ and $\abs{S_D} \leq 2\eta_0\abs{S}$.
\end{lemma}
All proofs for the theorem and lemmas in this section can be found in Appendix~\ref{sec:sample-comp-supp}.

\subsection{Lower bound of sample complexity}\label{subsec:lower-bound}

Though lower bound of $\Omega(s\log d)$ is already known for general learnability of $s$-sparse halfspaces in the distribution-free settings. However, for learning from fixed distributions under the Assumption~\ref{ass:dataset-margin} and \ref{ass:distribution-marginal}, it is unknown whether such bound still holds. Interestingly, \cite{shen2021attribute} showed that the same lower bound holds for any single isotropic logconcave marginals. Here, we generalize the result to any mixture of logconcaves even under the margin conditions and derive Lemma~\ref{lem:lower-bound}. 
%the question with affirmation by proving Lemma~\ref{lem:lower-bound}.

It is worth noting that for prior works on robust learning halfspaces depending on isotropic logconcave assumptions, a key property that was significantly used is that bounding the angle between two vectors $u,v$ implies bounding the disagreement probability between halfspace induced by $u,v$. This is due to that the probability density is uniform over all directions. This results in a key restriction for these algorithms that they only works for homogeneous halfspaces. Now, consider $\D_X,\D$ satisfying conditions in Assumption~\ref{ass:dataset-margin} with $\gamma\leq\frac12$and \ref{ass:distribution-marginal}, apparently, the probability density of such underlying distribution is no longer uniform over all directions $u,v$. As a result, we construct a hardness example, where the distribution density is uniform around $w^*$, and $0$ inside the margin. We show that, the volume of the region $A'$ where the probability density is non-zero is large enough, such that any two $u,v$ with disagreement region falling inside $A'$ will have their $L_2$ distance, i.e. $\twonorm{u-v}\geq c\cdot\epsilon$, implying a significant disagreement probability, i.e. $\Pr_{x\sim \D_X} \( \sign (u \cdot x) \neq \sign (v \cdot x) \) \geq \epsilon$. The complete proof is deferred to Appendix~\ref{sec:lower-bound-app}.

\subsection{Proof of Theorem~\ref{thm:main}}\label{subsec:proof-sketch}

The proof idea of our main theorem is to utilize the key deterministic result (Theorem~\ref{thm:deterministic-restate}) and show that by running Algorithm~\ref{alg:main} on a large enough sample set, the returned $\hat{w}$ has low error rate with respect to the underlying distribution $\D$.

\begin{proposition}\label{prop:feasibility}
Given that Asssumption~\ref{ass:distribution-marginal} holds, if we draw a sample set $S$ from $\EX$ with size $\abs{S}\geq \Omega(s^2\cdot \log^5d\log^2\frac{1}{\delta'}+\frac{1}{\eta_0}\cdot\log\frac{1}{\delta'})$, with probability at least $1-2\delta'$, the semidefinite program in Algorithm~\ref{alg:soft} is feasible. 
%Namely, Assumption~\ref{ass:feasibility} is satisfied.
\end{proposition}

%Assumption~\ref{ass:feasibility}
Proposition~\ref{prop:feasibility} shows that Program~\eqref{eq:hinge-loss-minimization} is feasible under a proper selection of the sample size. The proof is in Appendix~\ref{sec:sample-comp-supp}. We are now ready to prove our main theorem.

\begin{proof}[Proof of Theorem~\ref{thm:main}]
%%Since $S_C$ is drawn from $\D$, by Assumption~\ref{ass:distribution-margin}, Assumption~\ref{ass:dataset-margin} is also satisfied. 
%Assumption~\ref{ass:dataset-margin} is implied by Assumption~\ref{ass:distribution-margin}.
%
%We then show that the semidefinite program in Algorithm~\ref{alg:soft} is feasible (Assumption~\ref{ass:feasibility}). That is, we show the existence of a $q^*$ that satisfies all three constraints. From Lemma~\ref{lem:clean-sample-complexity}, we know that $\abs{S_C} \geq (1-2\eta_0)\abs{S}$ with probability $1-\delta'$, if $\abs{S}\geq\frac{3}{\eta_0}\cdot\log\frac{1}{\delta'}$. From Lemma~\ref{lem:clean-variance} and \ref{lem:clean-sample-complexity}, we know that if $\abs{S_C}\geq\Omega\(s^2\cdot \log^5(nd)\log^2\frac{1}{\delta'}\)$, it holds with probability $1-\delta'$,
%\begin{equation*}
%	\sup_{H:H\succeq0,\onenorm{H}\leq s,\nuclearnorm{H}\leq1} \frac{1}{\abs{S_C}}\sum_{i\in S_C} q_i x_i^{\top}Hx_i \leq 4\(\frac{1}{d}+r^2\) = \bar{\sigma}^2 .
%\end{equation*}
%Let $\eta_0\leq\frac14$, we have that $\abs{S_C}\geq \frac{1}{2}\abs{S}$. The above conditions are satisfied with probability at least $1-2\delta'$ if we choose $\abs{S}\geq \Omega\(s^2\cdot \log^5(nd)\log^2\frac{1}{\delta'}+\frac{1}{\eta_0}\cdot\log\frac{1}{\delta'}\)$.
%Hence, by setting $q^*_i=1$ for every $i\in S_C$ and setting all other weights to $0$, $q^*$ satisfies all three constraints. We have showed that the program is feasible.
From Proposition~\ref{prop:feasibility}, we know that the semidefinite program in Algorithm~\ref{alg:soft} is feasible with probability $1-2\delta'$, given that $\abs{S}\geq \Omega(s^2\cdot \log^5d\log^2\frac{1}{\delta'}+\frac{1}{\eta_0}\cdot\log\frac{1}{\delta'})$. 
From Theorem~\ref{thm:pancake-sample-complexity}, we obtain that with probability at least $1-\delta'$, $(S_C,\D)$ satisfies $\(\tau,\rho,\epsilon\)$-dense pancake condition by setting $\abs{S_C}\geq\Omega\(\frac{4k}{1-\epsilon}\cdot\({s}\cdot \log^4{d} +\log\frac{1}{\delta'\epsilon}\)\)$ with $\tau=2\sigma\cdot\(\log\frac{1}{\epsilon}+1\)$. 

Given that Assumption~\ref{ass:dataset-margin} holds with $\gamma>0$,
$(S_C, \D)$ satisfies $(\tau,\rho,\epsilon)$-dense pancake condition with $\tau\leq\frac{\gamma}{2}$, 
%and \ref{ass:pancake-deterministic} hold with $\tau = \frac{4(\log\frac{1}{\epsilon} +1)}{\sqrt{d}}$, 
and Program~\eqref{eq:hinge-loss-minimization} is feasible, it suffices to choose $\gamma\geq 2\tau =  \frac{8(\log\frac{1}{\epsilon} +1)}{\sqrt{d}}$ to satisfy the conditions for Theorem~\ref{thm:deterministic-restate}. As a result, we have $\rho\geq {16\(\frac{1}{\gamma\sqrt{d}}+\frac{r}{\gamma}+1\)\sqrt{\eta_0}}$
%Equation~\eqref{eq:rho-setting} holds 
when $k\leq 64$ and $\eta_0\leq\frac{1}{2^{28}}$. By setting $\delta'=\delta/3$, it suffices to choose 
%$\Omega\(s^2\log^5{d}\cdot\log^2\frac{1}{\delta} + s\log^4 d + \log\frac{1}{\delta\epsilon}\)$ 
$\Omega\(s^2\log^5{d}\cdot\log^2\frac{1}{\delta\epsilon}\)$. The proof is complete.
%Note that  sample complexity in Theorem~\ref{thm:main} satisfies this condition.
\end{proof}
%It is worth noting that the sample complexity of the semidefinite program in Algorithm~\ref{alg:soft} is significantly larger that that of the constrained hinge loss minimization program. In other words, if one was able to avoid the soft outlier removal, a better sample complexity may be achieved.

\section{Conclusion}

In this paper, we revisit the problem of attribute-efficient PAC learning of sparse halfspaces under the challenging malicious noise model, with a noise rate up to a constant. To the best of our knowledge, this is the first attribute-efficient algorithm that works under a  malicious noise rate of $\omega(\epsilon)$. It will be interesting to study if our gradient analysis can be extended to other learning settings, e.g. other surrogate loss functions. It is also interesting to apply our adaptation of single logconcave to mixture of logconcaves analysis in other problems, e.g. active learning under mixture of logconcaves.
%e.g. online learning; other surrogate loss functions; or other learning problems, e.g. multiclass classification.

%\section*{Impact Statement}
%
%This paper presents work whose goal is to advance the field of Machine
%Learning. There are many potential societal consequences of our work, none
%which we feel must be specifically highlighted here.

%\clearpage
\bibliographystyle{alpha}
\bibliography{../../../szeng_ref}

\clearpage
\appendix
\section{Omitted Proofs}

%Consider Algorithm~\ref{alg:main} and \ref{alg:soft}.

%\subsection{Ommitted Proofs in Section~\ref{sec:alg}}\label{subsec:alg-analysis}

We use $C$ with subscripts to denote absolute constants.

\subsection{Gradient and subgradient analysis of Algorithm~\ref{alg:main} (omitted proofs in Section~\ref{subsec:subgradient})}\label{subsec:subgradient-supp}

%Without the loss of generality, from now on, we fix the parameter $\gamma$ and omit the subscript $\gamma$ in loss function $\ell(\cdot)$.

Note that since we follow the theoretical framework of~\cite{shen2025efficient}, some proof structure might be similar. However, we study the problem in the attribute-efficient learning setting and  definitions have shifted. We include most of detailed proofs for completeness.

Recall that we have decomposed the hinge loss into different parts to facilitate our analysis. That is, for any given $(x,y)$, we consider that the set $S$ consists of three parts, i.e. $S_P, S_{\bar{P}}, S_D$. 
%(see Eq.~\eqref{eq:S-union}). 
Similarly, their contribution to the hinge loss function can be decomposed in the following way:
\begin{equation*}
\ell(w;q\circ S) = \ell(w;q\circ S_P) + \ell(w;q\circ S_{\bar{P}}) + \ell(w;q\circ S_D).
\end{equation*}
We summarize some characteristics for the function gradients. Namely, for any data set $S$ and weight vector $q$, the weighted subgradient is given by $\partial_w \ell(w;q\circ S) = \frac{1}{\gamma}\cdot \sum_{i\in S} \partial f(y_ix_i\cdot \frac{w}{\gamma})\cdot y_i(q_ix_i)$ where $f(z):=\max\{0,1-z\}$
% (see Eq.~\eqref{eq:normalized-hinge}). 
 It is worth noting that $\forall z$, $\partial f(z) \subseteq [-1,0]$.

We consider Algorithm~\ref{alg:main} for the rest of this section.
The goal is to show that, if some significant $(x,y)$ is misclassified by $\hat{w}$, a large number of good samples around $(x,y)$ will push the program to further optimize over $\hat{w}$. That is, the clean samples in set $S_P$ will have non-zero subgradients that contribute significantly to the updating step. 
We start with gradients from samples in $S_P$.
%\Shiwei{This is a reproduction of Lemma~14 in~\cite{shen2025efficient} by replacing the normalization with $\ell_1$ norm.}
\begin{lemma}\label{lem:subgradient-pancake}
	Consider a sample $(x,y)$ and its pancake $\P_{\hat{w}}^{\tau}(x,y)$ with $\tau\leq\gamma/2$. If $(x,y)$ is misclassified by $\hat{w}$, i.e. $yx\cdot\hat{w}\leq0$, then $\forall i\in S_P$, we have that $y_ix_i\cdot\hat{w}<\gamma$ and $\partial f(y_ix_i\cdot\frac{\hat{w}}{\gamma})=\{-1\}$.
\end{lemma}
\begin{proof}
Since $i\in S_P$, according to the dense pancake condition (Definition~\ref{def:pancake}), we have that $y_ix_i\cdot \hat{w} \leq yx\cdot \hat{w} + \tau$. Since $yx\cdot \hat{w} \leq0$ and $\tau\leq\gamma/2 < \gamma$, we have that $y_ix_i\cdot \hat{w} < \gamma$. We conclude that $\partial f(y_ix_i\cdot \frac{\hat{w}}{\gamma})=\{-1\}$.
%When $\hat{v}$ is returned from the program, it holds that $\onenorm{\hat{v}} \leq 1/\gamma$, and since $\hat{w} = \hat{v}/\onenorm{\hat{v}}$, we have that $y_ix_i\cdot \hat{v} < 1$, implying that $\partial f(y_ix_i\cdot\hat{v})=\{-1\}$.
\end{proof}
The following lemma connects the gradient norm (Definition~\ref{def:linsum}) with the actual subgradients from malicious samples, i.e. $S_D$. Specifically, the gradient norm serves as an upper bound for their weighted subgradients projecting onto any $w\in\W$.
%The following lemma upper bounds the influence of the weighted subgradients from the malicious samples with their gradient norm .
%The following lemma plays an essential role in our proof in the sense that it upper bounds the influence of the weighted subgradients from the malicious samples with their linear sum $\ell_\infty$ norm (Definition~\ref{def:linsum}).
\begin{lemma}\label{lem:subgradient-malicious}
For any vector $w\in\W$, 
%with $\onenorm{w}={\color{red}\sqrt{s}}$ and $\twonorm{w}\leq1$, 
given any $g\in\partial_w \ell(w;q\circ S_D)$, it holds that $g\cdot w \leq \frac{1}{\gamma}\cdot\linsum(q\circ S_D)$.
\end{lemma}
\begin{proof}
For $S_D$, we have that $\partial_w \ell(w;q\circ S_D) = \frac{1}{\gamma}\cdot \sum_{i\in S_D} \partial f(y_ix_i\cdot \frac{w}{\gamma})\cdot y_i(q_ix_i)$. Since $\partial f(y_ix_i\cdot \frac{w}{\gamma}) \subseteq [-1,0], \forall i$, we have that
\begin{align*}
g\cdot w &\leq \frac{1}{\gamma}\cdot \sup_{a_i\in[-1,0]} \inner{\sum_{i\in S_D}a_i y_i(q_ix_i)}{w}  \\
&\leq \frac{1}{\gamma}\cdot \sup_{a_i\in[-1,1]} \sup_{w\in\W} \inner{\sum_{i\in S_D} a_i q_ix_i}{w} \\
&= \frac{1}{\gamma}\cdot \linsum(q\circ S_D)
\end{align*}
due to Definition~\ref{def:linsum}.
%\Shiwei{Note that here we restrict $a_i$ in either $[0,1]$ or $[-1,0]$ would be the same effect.}
\end{proof}
The following lemma contributes to the analysis of subgradients $g\in\partial_w \ell(w;q\circ S)\big|_{w=\hat{w}}$ in the direction of $w^*$.
\begin{lemma}\label{lem:subgrad-star}
Assume that Assumption~\ref{ass:dataset-margin} holds. Given any $(x,y)$ and its pancake $\P_{\hat{w}}^{\tau}(x,y)$ with $\tau\leq \gamma/2$, if $(x,y)$ is misclassified by $\hat{w}$, i.e. $yx\cdot\hat{w} \leq 0$, then $\forall g\in \partial_w \ell(w;q\circ S) \big|_{w=\hat{w}}$, we have
\begin{equation*}
g\cdot w^* \leq -\sum_{i\in S_P} q_i + \frac{1}{\gamma}\cdot \linsum(q\circ S_D)
\end{equation*}
\end{lemma}
\begin{proof}
This lemma can be proved in three steps.
%and follows similar ideas from that of~\cite{shen2025efficient}. Since we are in a different setting and some definitions are shifted from the prior work, we include this proof for completeness. 
Namely, we will show that
\begin{enumerate}
\item For any $g_1\in \partial_w \ell(w;q\circ S_P)\big|_{w=\hat{w}}$, it holds that $g_1\cdot w^* \leq -\sum_{i\in S_P} q_i$;
\item For any $g_2\in \partial_w \ell(w;q\circ S_{\bar{P}})\big|_{w=\hat{w}}$, it holds that $g_2\cdot w^* \leq 0$;
\item For any $g_3\in \partial_w \ell(w;q\circ S_D)\big|_{w=\hat{w}}$, it holds that $g_3\cdot w^* \leq \frac{1}{\gamma}\cdot \linsum(q\circ S_D)$.
\end{enumerate}

To show the first statement, it is known from Lemma~\ref{lem:subgradient-pancake} that $\forall i\in S_P$, it holds that $\partial f(y_ix_i\cdot \frac{w}{\gamma})  = \{-1\}$. Hence, $g_1\cdot w^* = -\frac{1}{\gamma}\cdot \sum_{i\in S_P} q_i y_ix_i \cdot w^*$. From Assumption~\ref{ass:dataset-margin}, we know that $y_ix_i\cdot w^* \geq \gamma$. In conclusion, $g_1\cdot w^* \leq - \sum_{i\in S_P} q_i$.

For the second part, note that $\forall i\in S_{\bar{P}}$, we still have $y_ix_i\cdot w^* \geq \gamma$ due to Assumption~\ref{ass:dataset-margin}. Hence, $g_2\cdot w^* \leq \frac{1}{\gamma}\cdot \sum_{i\in S_{\bar{P}}} \partial f(y_ix_i\cdot\frac{w}{\gamma}) \cdot q_i (y_ix_i \cdot w^*) \leq 0$ since $\partial f(\cdot)$ is always non-positive.

For the last statement, it follows from Lemma~\ref{lem:subgradient-malicious} with $w^*\in \W$.
%the output condition of $\hat{w}$ of Algorithm~\ref{alg:main}.
%The proof, given that Assumption~\ref{ass:dataset-margin}, Lemma~\ref{lem:subgradient-pancake} and \ref{lem:subgradient-malicious} hold, follows from that of Lemma~18 in~\cite{shen2025efficient}.
\end{proof}

The next step is to show that the (negative) weighted subgradients, i.e. $- \partial_w \ell(w;q\circ S) \big|_{w=\hat{w}}$, is always aligned with $w^*-\hat{w}$. This is enabled by some well-defined $w'$, which is concluded in Lemma~\ref{lem:subgrad-prime}, and restated as follows.

%%%%%%%%%%%Old design of $w'$, was miscalculated.
%\Shiwei{Define a new vector $w' = \frac{w^*-\hat{w}\inner{w^*}{\beta}}{\norm{\cdot}}$, where $\beta = \frac{1-\lambda_2}{\kappa\lambda_1}\cdot\hat{z} + \frac{1-\kappa\lambda_1}{\lambda_2}\cdot\hat{w}$ and $\kappa=\sqrt{s}$ is the upper bound for $\onenorm{w}$. We will show that when the program output $\hat{w}$, and if $\hat{w}$ is on the boundary of both constraint, i.e. $g=-(\lambda_1\hat{z}+\lambda_2\hat{w})$, $\lambda_1,\lambda_2 >0$, $g\cdot w'=0$.}

\begin{lemma}[Restatment of Lemma~\ref{lem:subgrad-prime}]\label{lem:subgrad-prime-restate}
Consider Algorithm~\ref{alg:main}. Suppose Assumption~\ref{ass:dataset-margin} holds. For some $\hat{z}\in \frac{\partial \onenorm{w}}{\partial w}\big|_{w=\hat{w}}$, define  $w' = w^*-\hat{w}\inner{w^*}{\kappa}$, where $\kappa = \frac{\lambda_1}{{\sqrt{s}}\lambda_1+\lambda_2} \cdot\hat{z} + \frac{\lambda_2}{{\sqrt{s}}\lambda_1+\lambda_2} \cdot\hat{w}$. Assume that $\inner{w^*}{\kappa}>0$. Then, we have the following holds. For any $(x,y)\in\X\times\Y$ and its pancake $\P_{\hat{w}}^{\tau}(x,y)$ with $\tau\leq\gamma/2$, if $(x,y)$ is misclassified by $\hat{w}$, then for any $g\in\partial_{w}\ell(w; q\circ S)\big|_{w=\hat{w}}$, we have
\begin{equation*}
	g\cdot w' \leq -\frac12 \sum_{i\in S_P}q_i + {\frac{2}{\gamma} \cdot} \linsum(q\circ S_D). 
\end{equation*}
\end{lemma}
\begin{proof}
We will show the inequality in three steps:
\begin{enumerate}
\item For any $g_1 \in \partial_w \ell(w;q\circ S_P) \big|_{w=\hat{w}}$, it holds that $g_1 \cdot w' \leq -\frac12 \sum_{i\in S_P} q_i$.
\item For any $g_2 \in \partial_w \ell(w;q\circ S_{\bar{P}}) \big|_{w=\hat{w}}$, it holds that $g_2 \cdot w' \leq 0$.
\item For any $g_3 \in \partial_w \ell(w;q\circ S_D) \big|_{w=\hat{w}}$, it holds that $g_3 \cdot w' \leq {\frac{2}{\gamma} \cdot} \linsum(q\circ S_D)$. 
\end{enumerate}
Here, we give a general proof for the case when $\lambda_1,\lambda_2 \neq 0$. When either $\lambda_1,\lambda_2$ equals to zero, the proof naturally follows.

To show the first statement, we consider $S_P$. Given any $\hat{w}$, we have that $\forall i$,
\begin{equation*}
y_ix_i\cdot w' = y_ix_i\cdot (w^*-\hat{w}\inner{w^*}{\kappa})
%\frac{w^*-\hat{w}\inner{w^*}{\kappa}}{\norm{w^*-\hat{w}\inner{w^*}{\kappa}}_{2}}
= y_ix_i\cdot w^* - y_ix_i\cdot\hat{w}\inner{w^*}{\kappa}
%\frac{y_ix_i\cdot w^* - y_ix_i\cdot\hat{w}\inner{w^*}{\kappa}}{\norm{w^*-\hat{w}\inner{w^*}{\kappa}}_{2}}.
\end{equation*}
The goal is to lower bound the above term with a positive multiplicative factor of $\gamma$.
%We first bound the numerator.
By Assumption~\ref{ass:dataset-margin}, $\forall i\in S_P$, we have that $y_ix_i\cdot w^* \geq \gamma$. In addition, it holds that $y_ix_i\cdot\hat{w} \leq yx\cdot\hat{w} + \tau \leq 0+\frac{\gamma}{2} = \frac{\gamma}{2}$ due to Definition~\ref{def:pancake} and pancake parameters. Hence, it remains to bound quantity $\inner{w^*}{\kappa}$. This is easy from the observation that when the program outputs $\hat{w}$, it holds that $\onenorm{\hat{w}}\leq \sqrt{s}$ and $\twonorm{\hat{w}}\leq1$. Therefore, $\inner{w^*}{\kappa} = \frac{\lambda_1}{{\sqrt{s}}\lambda_1+\lambda_2} \cdot \inner{w^*}{\hat{z}} + \frac{\lambda_2}{{\sqrt{s}}\lambda_1+\lambda_2} \cdot \inner{w^*}{\hat{w}} \leq \frac{\lambda_1 \sqrt{s}}{{\sqrt{s}}\lambda_1+\lambda_2} + \frac{\lambda_2}{{\sqrt{s}}\lambda_1+\lambda_2} \leq1$, where the second transition is due to that $\inner{w^*}{\hat{z}} \leq \sup_{\infnorm{z}\leq1}\inner{w^*}{z}=\onenorm{w^*}=1$ and $\inner{w^*}{\hat{w}} \leq \sup_{\twonorm{w}\leq1}\inner{w^*}{w}=\twonorm{w^*}=1$.
%For the denominator, it is easy to show that $\twonorm{w^*-\hat{w}\inner{w^*}{\kappa}} \leq \twonorm{w^*}+\twonorm{\hat{w}\inner{w^*}{\kappa}} \leq 2$.
Hence, we have that $\forall i, y_ix_i\cdot w' \geq \gamma-\frac{\gamma}{2} \geq \frac{\gamma}{2}$.

%%%%%The following was the original proof for the case when there is only an $\ell_1$ constraint. - START
%\begin{equation*}
%y_ix_i\cdot\hat{w} \leq yx\cdot\hat{w} + \tau \leq 0+\frac{\gamma}{2} = \frac{\gamma}{2}.
%\end{equation*}
%Given any $\hat{z}$, by the definition of $w'$ we have
%\begin{equation*}
%	y_ix_i\cdot w' = \frac{y_ix_i\cdot w^* - y_ix_i\cdot\hat{w}\inner{w^*}{\hat{z}}}{\onenorm{w^* - \hat{w}\inner{w^*}{\hat{z}}}}.
%\end{equation*}
%For the numerator, since $\sup_{\infnorm{\hat{z}}\leq1} \inner{w^*}{\hat{z}} = \onenorm{w^*}=1$, we have that $y_ix_i\cdot w^* - y_ix_i\cdot\hat{w}\inner{w^*}{\hat{z}} \geq \gamma - \frac{\gamma}{2}\cdot\inner{w^*}{\hat{z}} = \gamma(1-\frac12\inner{w^*}{\hat{z}}) \geq \frac{\gamma}{2}$. For the denominator, $\onenorm{w^* - \hat{w}\inner{w^*}{\hat{z}}} \leq \onenorm{w^*} + \onenorm{\hat{w}\inner{w^*}{\hat{z}}} \leq2$. Thus, $y_ix_i\cdot w' \geq \frac{\gamma}{4}$. 
%%%%%The above was the original proof for the case when there is only an $\ell_1$ constraint. - END

On the other hand, Lemma~\ref{lem:subgradient-pancake} imlies that $\partial f(y_ix_i\cdot\frac{w}{\gamma})=\{-1\}$ for $i\in S_P$. Hence,
\begin{align*}
\partial_w \ell(w;q\circ S_P) \big|_{w=\hat{w}} \cdot w' &= \frac{1}{\gamma}\cdot \sum_{i\in S_P}q_i \partial f\(y_ix_i\cdot\frac{w}{\gamma}\)y_ix_i\cdot w' \\
&\subseteq \left(-\infty,-\frac12\sum_{i\in S_P}q_i\right].
\end{align*}

Then, we consider $i\in S_{\bar{P}} \subseteq S_C$. If $y_ix_i\cdot w' \geq 0$, then 
\begin{align}
\partial_w \ell(w;q\circ S_{\bar{P}}) \big|_{w=\hat{w}} \cdot w' &= \frac{1}{\gamma}\cdot \sum_{i\in S_{\bar{P}}}q_i \partial f\(y_ix_i\cdot\frac{w}{\gamma}\) y_ix_i\cdot w' \notag \\
&\subseteq (-\infty,0], \label{eq:condi-1}
\end{align}
since $\partial f(\cdot)$ is non-positive. We then consider $y_ix_i\cdot w' < 0$, namely,
\begin{equation*}
y_ix_i\cdot w' = y_ix_i\cdot w^* - y_ix_i\cdot\hat{w}\inner{w^*}{\kappa}
%\frac{y_ix_i\cdot w^* - y_ix_i\cdot\hat{w}\inner{w^*}{\kappa}}{\twonorm{w^* - \hat{w}\inner{w^*}{\kappa}}} 
<0.
\end{equation*}
By assumption, $\inner{w^*}{\kappa} >0$ (see Lemma~\ref{lem:sign-of-w-kappa} for conditions for it to hold). Then, we have
\begin{equation*}
y_ix_i\cdot\hat{w} > \frac{y_ix_i\cdot w^*}{\inner{w^*}{\kappa}} \geq \gamma,
\end{equation*}
where the last inequality is due to Assumption~\ref{ass:dataset-margin}.
Rescaling by a factor of $\frac{1}{\gamma}$, we have that $y_ix_i\cdot\frac{\hat{w}}{\gamma}>1$. Thus,

\begin{equation*}
\partial_w \ell(w;q\circ S_{\bar{P}}) \big|_{w=\hat{w}} \cdot w' = \frac{1}{\gamma}\cdot \sum_{i\in S_{\bar{P}}}q_i\partial f\(y_ix_i\cdot\frac{\hat{w}}{\gamma}\) y_ix_i\cdot w' = \{0\}.
\end{equation*}
Together with Eq.\eqref{eq:condi-1}, we conclude that the second part holds. 

The third part follows directly from Lemma~\ref{lem:subgradient-malicious} by concluding that $\onenorm{w'} = \onenorm{w^*-\hat{w}\inner{w^*}{\kappa}} \leq \onenorm{w^*} + \onenorm{\hat{w}}\cdot\inner{w^*}{\kappa} \leq 2\sqrt{s}$ and $\twonorm{w'}\leq2$. Namely, $\frac{w'}{2}\in\W$.
%$\onenorm{w'}= \onenorm{ \frac{w^*-\hat{w}\inner{w^*}{\kappa}}{\twonorm{w^*-\hat{w}\inner{w^*}{\kappa}}} } \leq ?  \leq 2\sqrt{s}$ 
%\Shiwei{this needs to be seriously considered, the intuition is that the rescaling by deviding the L2 norm should not change the L1 norm too much}. \Shiwei{2025.04.10: I stuck here...}
The proof is complete. 
%the $\ell_1$ norm of $w'$ by ${\color{red}\sqrt{s}}$.
\end{proof}

\begin{lemma}[Restatement of Lemma~\ref{lem:choice-of-w'}]\label{lem:choice-of-w'-restate}
Consider the set of subgradients $\partial_{w}\ell(w; q\circ S)\big|_{w=\hat{w}}$ and some element $g$ in it. Given the optimization program Eq.\eqref{eq:hinge-loss-minimization}  in Algorithm~\ref{alg:main} and $\hat{w}$ returned by it, let $w' = w^*-\hat{w}\inner{w^*}{\kappa}$. By choosing $\kappa = \frac{\lambda_1}{{\sqrt{s}}\lambda_1+\lambda_2} \cdot\hat{z} + \frac{\lambda_2}{{\sqrt{s}}\lambda_1+\lambda_2} \cdot\hat{w}$ for some $\lambda_1,\lambda_2\geq0, \lambda_1+\lambda_2>0$ and $\hat{z}\in\partial_w\onenorm{w}\big|_{w=\hat{w}}$, there exists $g\in \partial_{w}\ell(w; q\circ S)\big|_{w=\hat{w}}$ such that $g\cdot w'=0$.
\end{lemma}
\begin{proof}
%We first show that $\exists g \in \partial_{w}\ell(w; q\circ S)\big|_{w=\hat{w}}$ such that $g\cdot w'=0$. 
The design of $w'$ stems from the program Eq.\eqref{eq:hinge-loss-minimization}, which we restate as follows
\begin{eqnarray*}
&\quad\ \underset{}{\arg\min}\ \ell (w;q \circ S)\\
&\text{s.t.} \onenorm{w}\leq {\sqrt{s}}, \\
&\ \twonorm{w}\leq1.
\end{eqnarray*}
Intuitively, $w'$ is the component of $w^*-\hat{w}$ that is orthogonal to some subgradient of the weighted hinge loss. %the constraints. 
From the KKT condition, we know that for any output $\hat{w}$ of the program, there exist some $g$ and $\hat{z}$ such that 
%\Shiwei{check $\exists/\forall$}
\begin{equation*}
g + \lambda_1\hat{z} + \lambda_2\hat{w} = 0
\end{equation*}
for some $\lambda_1,\lambda_2\geq0$. Taking this $g$ and $\hat{z}$,  consider the design of $w'$, we have that
\begin{equation*}
g\cdot w' = -(\lambda_1\hat{z}+\lambda_2\hat{w}) \cdot (w^*-\hat{w}\inner{w^*}{\kappa})
%\frac{w^*-\hat{w}\inner{w^*}{\beta}}{\twonorm{w^*-\hat{w}\inner{w^*}{\beta}}} 
%&= -(\lambda_1\hat{z}+\lambda_2\hat{w}) \cdot \frac{w^*-\hat{w}\inner{w^*}{\frac{\lambda_1}{{\sqrt{s}}\lambda_1+\lambda_2} \cdot\hat{z} + \frac{\lambda_2}{{\sqrt{s}}\lambda_1+\lambda_2} \cdot\hat{w}}}{\twonorm{w^*-\hat{w}\inner{w^*}{\kappa}}} \\ 
\end{equation*}
To show $g\cdot w'=0$, it suffices to show that
%To show that $g\cdot w'$ equals zero, it suffices to show that $-(\lambda_1\hat{z}+\lambda_2\hat{w}) \cdot \(w^*-\hat{w}\inner{w^*}{\kappa}\) = 0$. That is,
\begin{align*}
-(\lambda_1\hat{z}+\lambda_2\hat{w})\cdot w^* +  (\lambda_1\hat{z}+\lambda_2\hat{w}) \cdot \hat{w}\inner{w^*}{\kappa} &= 0, \\
(\lambda_1\hat{z}+\lambda_2\hat{w}) \cdot \hat{w}\inner{w^*}{\kappa} &=  (\lambda_1\hat{z}+\lambda_2\hat{w})\cdot w^*.
\end{align*}
Assuming that both $\lambda_1,\lambda_2 \neq 0$ (either one equals zero would be a special case and follows easily from this proof; both equal zero would lead to an interior), then by the complementary slackness, the boundary conditions are achieved, i.e. $\hat{z}\cdot\hat{w}=\onenorm{\hat{w}}=\sqrt{s}$ and $\hat{w}\cdot\hat{w}=\twonorm{\hat{w}}=1$. Hence, the equality to be shown becomes
\begin{align*}
(\lambda_1\sqrt{s}+\lambda_2)\inner{w^*}{\kappa} = \inner{\lambda_1\hat{z} + \lambda_2\hat{w}}{w^*}.
\end{align*}
It suffices to choose $\kappa = \frac{\lambda_1\hat{z} + \lambda_2\hat{w}}{\lambda_1\sqrt{s}+\lambda_2}$. The proof is complete.
\end{proof}

\begin{lemma}\label{lem:sign-of-w-kappa}
%Consider the set of subgradients $\partial_{w}\ell(w; q\circ S)\big|_{w=\hat{w}}$ and some element $g$ in it. 
Consider that the weight from $S_P$ is prevalent (Lemma~\ref{lem:weight-clean}).
Given program Eq.\eqref{eq:hinge-loss-minimization}  in Algorithm~\ref{alg:main} and $\hat{w}$ returned by it, let $w' = w^*-\hat{w}\inner{w^*}{\kappa}$ and $\kappa = \frac{\lambda_1}{{\sqrt{s}}\lambda_1+\lambda_2} \cdot\hat{z} + \frac{\lambda_2}{{\sqrt{s}}\lambda_1+\lambda_2} \cdot\hat{w}$ for some $\lambda_1,\lambda_2\geq0, \lambda_1+\lambda_2>0$ and $\hat{z}\in\partial_w\onenorm{w}\big|_{w=\hat{w}}$. Then, it holds that $\inner{w^*}{\kappa} >0$.
\end{lemma}
\begin{proof}
%We proceed to show the second statment.
According to Lemma~\ref{lem:weight-clean}, we have that 
\begin{equation*}
	\sum_{i\in S_P} q_i > \frac{1}{\gamma}\cdot \linsum(q\circ S_D).
\end{equation*}
Together with Lemma~\ref{lem:subgrad-star}, we conclude that  $\forall g$, $g\cdot w^* \leq -\sum_{i\in S_P} q_i + \frac{1}{\gamma}\cdot \linsum(q\circ S_D) <0$. Hence,
\begin{align*}
	(\lambda_1\sqrt{s}+\lambda_2)\inner{w^*}{\kappa} = \inner{\lambda_1\hat{z} + \lambda_2\hat{w}}{w^*} = -g\cdot w^* >0,
\end{align*}
implying that $\inner{w^*}{\kappa} > 0$ because $(\lambda_1\sqrt{s}+\lambda_2)>0$. The proof is complete.
\end{proof}

We are now ready to prove Theorem~\ref{thm:local-correctness}.
\begin{proof}[Proof of Theorem~\ref{thm:local-correctness}]
We will show by contradition. 
Let us assume that $(x,y)$ is misclassified by $\hat{w}$. 

{\bfseries Case 1.}\ Suppose that $\hat{w}$ is an interior of the constraint set of program~\eqref{eq:hinge-loss-minimization}. From Lemma~\ref{lem:subgrad-star}, we know that $\forall g \in \partial_w \ell(w;q\circ S)\mid_{w=\hat{w}}$, it holds that 
\begin{equation*}
	g\cdot w^* \leq -\sum_{i\in S_P} q_i + \frac{1}{\gamma}\cdot \linsum(q\circ S_D).
\end{equation*}
From Eq.~\eqref{eq:density-noise-rate}, we conclude that $\forall g, g\cdot w^*<0$. However, from the optimility condition  of program~\eqref{eq:hinge-loss-minimization}, there exists some $g=0$ such that $g\cdot w^*=0$, leading to a contradition.

{\bfseries Case 2.}\  Suppose that $\hat{w}$ is on the boundary of the constraint set. From Lemma~ \ref{lem:subgrad-prime}, we know that $\forall g$, it holds that 
\begin{equation*}
	g\cdot w' \leq -\frac12 \sum_{i=1}^{n}q_i + {\frac{2}{\gamma} \cdot} \linsum(q\circ S_D). 
\end{equation*}
Again from Eq.~\eqref{eq:density-noise-rate}, we conclude that $\forall g, g\cdot w'<0$. However, by the design of $w'$ and Lemma~\ref{lem:choice-of-w'}, there exists some $g$ such that $g\cdot w'=0$, leading to contradiction.

Hence, $(x,y)$ is not misclassfied by $\hat{w}$.
\end{proof}

\subsection{Analysis of density and noise rate conditions}\label{subsec:density-app}

From Section~\ref{subsec:subgradient}, it is evident that the essential condition for an $(x,y)$ to be correctly classified by the optimum $\hat{w}$ is that the weight of the clean samples around it is large enough comparing to the gradient norm from the malicious samples. This condition is given in Eq.~\eqref{eq:density-noise-rate}. The following lemma ensures that the condition is satisfied.
%The following lemma ensures that the conditions in Theorem~\ref{thm:local-correctness} are satisfied.
\begin{lemma}\label{lem:weight-clean}
Consider Algorithm~\ref{alg:soft} and its returned value $q$. Assume that Program~\eqref{eq:hinge-loss-minimization} is feasible. If $(x,y)$ is a sample such that the pancake $\P_{\hat{w}}^{\tau}(x,y)$ is $\rho$-dense with respect to $S_C$ with $\rho\geq {16\(\frac{1}{\gamma\sqrt{d}}+\frac{r}{\gamma}+1\)\sqrt{\eta_0}}$, then
it holds that 
\begin{equation}\label{eq:clean-prevalent}
%\frac14
\sum_{i\in S_C \cap \P_{\hat{w}}^{\tau}(x,y)} q_i > \frac{4}{\gamma}\cdot\linsum(q\circ S_D).
\end{equation}
\end{lemma}

%We provide a complete proof for Lemma~\ref{lem:weight-clean-restate}, which we restate as follows.

%\begin{lemma}[Restatement of Lemma~\ref{lem:weight-clean}]\label{lem:weight-clean-restate}
%	Consider Algorithm~\ref{alg:soft} and its returned value $q$. Assume that the program is feasible (Assumption~\ref{ass:feasibility}). If $(x,y)$ is a sample such that the pancake $\P_{\hat{w}}^{\tau}(x,y)$ is $\rho$-dense with respect to $S_C$ with $\rho\geq {16\(\frac{1}{\gamma\sqrt{d}}+\frac{r}{\gamma}+1\)\sqrt{\eta_0}}$, then
%	it holds that 
%	\begin{equation}\label{eq:clean-prevalent-restate}
%		\sum_{i\in S_C \cap \P_{\hat{w}}^{\tau}(x,y)} q_i > \frac{4}{\gamma}\cdot\linsum(q\circ S_D).
%	\end{equation}
%\end{lemma}
\begin{proof}%[Full proof of Lemma~\ref{lem:weight-clean}]
	We will prove the lemma by lower bounding the left hand side and upper bounding the right hand side respectively. 
	That is, we will show that $\sum_{i\in S_C\cap \P_{\hat{w}}^{\tau}(x,y)} q_i  = \sum_{i\in S_P} q_i  > (\rho - 2\xi)\abs{S}$; %Lemma 10 in non-sparse paper draft
	and $\linsum(q\circ S_D) \leq \bar{\sigma} \cdot\sqrt{\xi}\cdot\abs{S}$. %Lemma 11 in non-sparse paper draft
	Then, given these two conditions, Eq.~\eqref{eq:clean-prevalent} follows from that $\rho\geq {16\Big(\frac{1}{\gamma\sqrt{d}}+\frac{r}{\gamma}+1\Big)\sqrt{\eta_0}}$ and $\xi=2\eta_0$. 
	In more details, it remains to show that
	\begin{align*}
		(\rho-2\xi)\abs{S} &> \frac{4}{\gamma}\cdot \bar{\sigma}\cdot\sqrt{\xi}\cdot\abs{S}, \\
		\rho-2\xi &> \frac{4\bar{\sigma}}{\gamma}\cdot\sqrt{\xi}. 
	\end{align*}
	Since $\xi=2\eta_0$ and $\bar{\sigma} = 2\cdot \sqrt{\(\frac{1}{d}+r^2\)}$, it remains to show
	\begin{align*}
		\rho &> 4\eta_0 + \frac{8\sqrt{2}}{\gamma}\cdot\sqrt{\frac{1}{d}+r^2}\cdot\sqrt{\eta_0}.
	\end{align*}
	It suffices to choose any $\rho\geq {16\(\frac{1}{\gamma\sqrt{d}}+\frac{r}{\gamma}+1\)\sqrt{\eta_0}}$, due to that $(a+b) \geq \sqrt{a^2+b^2}$ for $a,b\geq0$ and $\sqrt{\eta_0}\geq\eta_0$.
	
	{\bfseries Lower bound.}\ 
	When Algorithm~\ref{alg:soft} outputs $q$, it holds that $\sum_{i\in S}q_i \geq (1-\xi)\abs{S}$. In addition, we have that $\abs{S_P}\geq\rho\abs{S_C}$ and $\abs{S_{\bar{P}}}\leq(1-\rho)\abs{S_C}$, given that the pancake $\P_{\hat{w}}^{\tau}(x,y)$ is $\rho$-dense w.r.t. $S_C $. 
	Since $S = S_P\cup S_{\bar{P}} \cup S_D$, it suffices to bound $\sum_{i\in S_{\bar{P}}}q_i \leq (1-\rho)\abs{S_C} \leq (1-\rho)\abs{S}$ and $\sum_{i\in S_D}q_i \leq \abs{S_D}\leq\xi\abs{S}$ where $\xi$ is the upper bound on empirical noise rate. Hence, $\sum_{i\in S_P} q_i = \sum_{i\in S} - \sum_{i\in S_{\bar{P}}}q_i - \sum_{S_D}q_i >(1-\xi-(1-\rho)-\xi)\abs{S} = (\rho-2\xi)\abs{S}$.
	
	{\bfseries Upper bound.}\ 
	Since that $q$ is feasible, we have
	\begin{equation*}
		\sum_{i\in S_D}q_i(w\cdot x_i)^2 \leq \sum_{i\in S} q_i\cdot(w\cdot x_i)^2 \leq \bar{\sigma}^2\cdot\abs{S}.
	\end{equation*}
	where the first transition is due to that all terms inside the summation is non-negative, and the second transition is due to Lemma~\ref{lem:variance-bound-restate}. 
	Therefore,
	\begin{align*}
		\sqrt{\sup_{w\in\W} \sum_{i\in S_D}q_i(w\cdot x_i)^2 } \leq \bar{\sigma}\sqrt{\abs{S}},
		%\sqrt{\sup_{\onenorm{w}\leq1} \sum_{i\in S_D}q_i(w\cdot x_i)^2 } \leq \frac{\bar{\sigma}}{\sqrt{s}}\sqrt{\abs{S}}.
	\end{align*}
	Then,
	\begin{align*}
		\linsum(q\circ S_D) &\leq \sqrt{\abs{S_D}} \cdot \sqrt{\sup_{w\in\W}\sum_{i\in S_D}q_i(w\cdot x_i)^2} \\
		%\sup_{a_i\in[-1,1],i\in S_D} \infnorm{ \sum_{i\in S_D}a_iq_ix_i } &\leq 
		&\leq \sqrt{\abs{S_D}} \cdot\bar{\sigma}\sqrt{\abs{S}} \\
		&\leq  \bar{\sigma} \cdot \sqrt{\xi} \cdot {\abs{S}}
	\end{align*}
	where the first transition follows from Lemma~\ref{lem:linsum-bound-restate}. 
	The proof is complete.
\end{proof}

\begin{lemma}%[Restatement of Lemma~\ref{lem:variance-bound}]
	\label{lem:variance-bound-restate}
	Consider Algorithm~\ref{alg:soft}. If it successfully returns a weight vector $q$, then for any $w\in\W$, it holds that $\frac{1}{\abs{S}}\sum_{i\in S} q_i\cdot(w\cdot x_i)^2 \leq \bar{\sigma}^2$.
\end{lemma}
\begin{proof}%[Proof of Lemma~\ref{lem:variance-bound}]
	By simple calculation, $\sum_{i\in S} q_i\cdot(w\cdot x_i)^2 = \sum_{i\in S} q_i\cdot x_i^{\top}ww^{\top} x_i = \sum_{i\in S} q_i\cdot x_i^{\top}W x_i$ where $W:=ww^{\top}$. 
	Since $\onenorm{w}\leq \sqrt{s}$, we have that $\onenorm{W}=\sum_{k,l\in[d]} \abs{W_{k,l}} = \sum_{k,l\in[d]} \abs{w_k \cdot w_l} \leq  \sum_{k\in[d]} \abs{w_k}\cdot\onenorm{w} \leq \onenorm{w}\cdot\onenorm{w} \leq s$.  Since $\twonorm{w}\leq1$, we have that $\nuclearnorm{W}\leq1$. Hence, $W\in\{H:H\preceq0,\onenorm{H}\leq s,\nuclearnorm{H}\leq1\}$, and $\sum_{i\in S} q_i\cdot(w\cdot x_i)^2 \leq \sup_{H\preceq0,\onenorm{H}\leq s,\nuclearnorm{H}\leq1} \frac{1}{\abs{S}}\sum_{i\in {S}} q_i x_i^{\top}Hx_i \leq \bar{\sigma}^2$. The proof is complete.
	%One the other hand, if the algorithm returns a $q$, then for any $\onenorm{H}\leq s$, it holds that $\frac{1}{\abs{S}}\sum_{i\in {S}} q_i x_i^{\top}Hx_i \leq \bar{\sigma}^2$. 
	%Hence, $\sum_{i\in S} q_i\cdot(w\cdot x_i)^2 \leq \sup_{\onenorm{H}\leq s} \frac{1}{\abs{S}}\sum_{i\in {S}} q_i x_i^{\top}Hx_i \leq \bar{\sigma}^2$. The proof is complete.
\end{proof}

\begin{lemma}%[Restatement of Lemma~\ref{lem:linsum-bound}]
	\label{lem:linsum-bound-restate}
	Let $S'$ be any sample set and let $q=(q_1,\dots,q_{\abs{S'}})$ with all $q_i\in[0,1]$. Then, $\linsum(q\circ S') \leq \sqrt{\abs{S'}}\sqrt{\sup_{w\in\W}\sum_{i\in S'}q_i(w\cdot x_i)^2}$.
\end{lemma}
\begin{proof}
	%Let us consider duel norm. 
	From the definition of $\linsum$ (Definition~\ref{def:linsum}), we have that %$\linsum(q\circ S') := \sup_{a_i\in[-1,1], i\in S'} \infnorm{\sum_{i\in S'} a_i q_i x_i}$.
	\begin{align*}
		\linsum(q\circ S') &= \sup_{a\in[-1,1]^{\abs{S'}}} \sup_{w\in \W} \inner{\sum_{i\in S'} a_i q_i x_i}{w} \\
		%&:= \sup_{a_i\in[-1,1], i\in S'} \infnorm{\sum_{i\in S'} a_i q_i x_i} \\
		&= \sup_{a_i\in[-1,1], i\in S'}\sup_{w\in \W} \inner{\sum_{i\in S'} a_i q_i x_i}{w} \\
		&= \sup_{a_i\in[-1,1], i\in S'}\sup_{w\in \W} \sum_{i\in S'} a_i q_i \inner{ x_i}{w} \\
		&\leq \sup_{a_i\in[-1,1], i\in S'}\sup_{w\in \W} \sqrt{\sum_{i\in S'} a_i^2} \sqrt{\sum_{i\in S'} (q_i \inner{ x_i}{w})^2 } \\
		&\leq \sqrt{S'} \sqrt{\sup_{w\in \W}  \sum_{i\in S'} q_i ({x_i}\cdot{w})^2 }
	\end{align*}
	%where the second transition is due to that $\ell_1$ and $\ell_\infty$ norms are duel to each other; 
	where the fourth transition is due to Cauchy-Schwartz inequality and the last transition is due to that $\forall i$, $a_i^2\leq1$ and $q_i^2\leq q_i$.
\end{proof}

\subsection{Analysis for deterministic results }\label{subsec:deterministic-app}

%We will prove the theorem in two steps. 
%We first show that if a given $(x,y)$ has a $\rho$-dense pancake $\P_{\hat{w}}^{\tau}(x,y)$ for  $\hat{w}$, and the clean samples in this pancake contribute to a sufficient weight,  then $(x,y)$ will not be misclassified by $\hat{w}$ (Theorem~\ref{thm:local-correctness}). 
%%and the parameter $\rho$ is large enough comparing to the noise rate
%We then show that, as long as $\rho$ is large enough comparing to $\eta_0$ (the upper bound on noise rate), when Algorithm~\ref{alg:soft} returns the weight $q$, the conditions for Theorem~\ref{thm:local-correctness} will be satisfied. This result is summarized in Lemma~\ref{lem:weight-clean} and  depends on our analysis for Algorithm~\ref{alg:soft}.
%% and a concentration result for logconcave distributions.
%%Last but not least, we conclude the guarantee of our main algorithm in Section~\ref{label}.

We note that this section follows directly from prior work of~\cite{shen2025efficient}. However, we must include it here as we have re-prove every theorems and lemmas in the new problem setting. See Section~\ref{sec:pre} for more details and we encourage interested readers to compare our problem setting with~\cite{shen2025efficient}.

\begin{theorem}%[Restatement of Theorem~\ref{thm:deterministic}]
\label{thm:deterministic-restate}
Given that Assumption~\ref{ass:dataset-margin} holds with some $\gamma>0$,
$(S_C, \D)$ satisfies $(\tau,\rho,\epsilon)$-dense pancake condition with $\tau\leq\frac{\gamma}{2}$, 
%and Assumption~\ref{ass:pancake-deterministic} holds with $\tau\leq\frac{\gamma}{2}$, 
the semidefinite program in Algorithm~\ref{alg:soft} is feasible, and %$\rho\geq {\color{red}16\(\frac{1}{\gamma\sqrt{d}}+\frac{r}{\gamma}+1\)\sqrt{\eta_0}}$, 
\begin{equation*}
	\rho\geq {16\(\frac{1}{\gamma\sqrt{d}}+\frac{r}{\gamma}+1\)\sqrt{\eta_0}}
\end{equation*}
then Algorithm~\ref{alg:main} returns $\hat{w}$ such that $\err_{D}(\hat{w}) \leq \epsilon$.
\end{theorem}
\begin{proof}%[Proof of Theorem~\ref{thm:deterministic}]
Since $(S_C, \D)$ satisfies $(\tau,\rho,\epsilon)$-dense pancake condition with $\tau\leq\frac{\gamma}{2}$ with respect to any $w$ with $\onenorm{w} \leq {\sqrt{s}}$ and $\twonorm{w}\leq1$ (Definition~\ref{def:pancake}). That is, for any unit vector $w$ such that $\onenorm{w}\leq\sqrt{s}$, $\Pr_{(x,y)\sim\D}(\P_{{w}}^\tau(x,y)\text{ is }\rho\text{-dense w.r.t. }S_C)\geq 1-\epsilon$.

Apparantly, this inequality holds for $\hat{w}$ as well due to the output condition in Eq.~\eqref{eq:hinge-loss-minimization}. Hence, for a $(1-\epsilon)$ probability mass in $\D$, $(x,y)$  has a $\rho$-dense pancake $\P_{\hat{w}}^\tau(x,y)$. Theorem~\ref{thm:local-correctness} together with Lemma~\ref{lem:weight-clean} implies that such $(x,y)$ will not be misclassified by $\hat{w}$. For the remaining $\epsilon$ probability mass, they might be misclassified by $\hat{w}$ and hence resulting in an error rate less or equal to $\epsilon$.
\end{proof}

\section{Analysis on Sample Complexity}\label{sec:sample-comp-supp}

%\subsection{Analysis for statistical results}

%\begin{proposition}\label{prop:sample-complexity}
%The sample complexity for Algorithm~\ref{alg:main} is 
%\begin{equation*}
%O\(s^2\cdot\log^5 d \cdot\log^2\frac{1}{\delta} + \frac{1}{\eta_0}\cdot\log\frac{1}{\delta}  +  \frac{k}{1-\beta}\(\frac{s}{(\epsilon')^2}\cdot \log\frac{d}{s} +\log\frac{1}{\delta\epsilon}\)     \).
%\end{equation*}
%\end{proposition}
%\begin{proof}
%The sample complexity comes from Lemma~\ref{lem:soft-sample-complexity}, Lemma~\ref{lem:clean-sample-complexity}, and Theorem~\ref{thm:pancake-sample-complexity} by setting $\delta'=\frac{\delta}{3}$.
%\end{proof}

%\Shiwei{We need to first show that the dense pancake condition holds for mixture of logconcaves with respect to all $\twonorm{w}=1, \onenorm{w}\leq\sqrt{s}$ vectors; then, since this holds, it also holds for all $\twonorm{w}\leq1$ and $\onenorm{w}\leq\sqrt{s}$.}
\begin{lemma}\label{lem:dense-distribution}
Consider a distribution $\D$ over $\X\times\Y$ satisfying Assumption~\ref{ass:distribution-marginal} and \ref{ass:dataset-margin} with $\gamma>\tau$ for some $\tau=2\sigma\cdot\(\log \frac{1}{\beta}+1\)$. Then, for any $\beta>0$, $\D$ satisfies $(\tau,\frac{1-\beta}{k},\beta)$-dense pancake condition. 
\end{lemma}
\begin{proof}
Consider a distribution $\D_1$ that is logconcave with zero-mean and covariance matrix $\Sigma\preceq\sigma^2 I_d$.  For any unit vector $w$, the projection of $\D_1$ onto  $w$ is still logconcave. The statistical property of logconcave random variable implies that
%For any $w$ such that $\onenorm{w}\leq\sqrt{s},\twonorm{w}=1$, 
\begin{equation*}
\Pr_{x\sim\D_1}\( \abs{w\cdot x} \leq\sigma\cdot\(\log \frac{1}{\beta}+1\)\) \geq 1-\beta.
\end{equation*}
It then immediately follows that, for any logconcave distribution $\D_j$ with mean $\mu_j$ and covariance matrix $\Sigma_j\prec\sigma^2 I_d$, we have 
\begin{equation*}
	\Pr_{x\sim\D_j}\(\abs{w\cdot (x-\mu_j)} \leq\sigma\cdot\(\log \frac{1}{\beta}+1\)\) \geq 1-\beta.
\end{equation*}
Note that this condition also holds for $w^*$. For any $x\sim\D_j$ such that $\abs{w^*\cdot(x-\mu_j)}\leq\sigma\cdot(\log1/\beta+1)=:\tau/2 < \gamma$, it holds that
%On the other hand, we want to show that with high probability, samples from the same distribution $\D_j$ have the same label by $w^*$. 
%This can be seen by the assumption of the $\gamma^*$-margin and the condition that the above inequality also holds for $w^*$. By definition, ${\color{red}2\sigma\cdot\(\log \frac{1}{\beta}+1\)  = \tau } < \gamma^*/2$, which implies that for any $x,x'$ that $\abs{w^*\cdot(x-x')}\leq$ 
$\sign(x\cdot w^*) =  \sign(\mu_j\cdot w^*)$.
%$\sign(x\cdot w^*) = \sign(\mu_j\cdot w^* \pm \tau/2) = \sign(\mu_j\cdot w^*)$ due to that $\tau<\gamma/2$. 

%% this is not necessary? at least, this can be shown with either large-margin assumption or separatable logconcaves assumption (\tau<\xi or \tau<\gamma) - Start 

%This can be seen by the assumption that the logconcave distributions are well seperated, i.e. $\abs{\mu_j\cdot w^*}\geq\xi$. By definition, ${\color{red}\sigma\cdot\(\log \frac{1}{\beta}+1\)  = \tau/2 } < \xi$, which implies that $\sign(x\cdot w^*) = \sign(\mu_j\cdot w^* \pm \tau/2) = \sign(\mu_j\cdot w^*)$. 
That is,
\begin{equation*}
	\Pr_{x\sim \D_j}\( \abs{\(\sign(x\cdot w^*)x-\sign(\mu_j\cdot w^*)\mu_j\)\cdot w} \leq {\tau}/{2} \) \geq 1-\beta.
\end{equation*}
%% this is not necessary? at least, this can be shown with either large-margin assumption or separatable logconcaves assumption - End
In addition, denoted by $(\D_j,w^*)$ the distribution where the instances is drawn from $\D$ and labeled by $w^*$, then
\begin{equation*}
	\Pr_{(x,y)\sim (\D_j,w^*)}\( \Pr_{(x',y')\sim (\D_j,w^*)} \( \abs{\(y'x'-yx\)\cdot w}\leq \tau \) \geq 1-\beta\) \geq 1-\beta.
\end{equation*}
Since $\DX=\sum_{j=1}^{k} \frac{1}{k}\D_j$, each distribution density of $\D_j$ is attenuated by a multiplicative factor of $\frac{1}{k}$. Hence, we have that
\begin{equation*}
	\Pr_{(x,y)\sim (\frac{1}{k}\D_j,w^*)}\( \Pr_{(x',y')\sim (\frac{1}{k}\D_j,w^*)} \(\abs{\(y'x'-yx\)\cdot w}\leq \tau \) \geq \frac{1-\beta}{k}\) \geq \frac{1-\beta}{k}.
\end{equation*}
%or, in other words,
%\begin{equation*}
%\Pr_{(x,y)\sim (\frac{1}{k}\D_j,w^*)}\( \Pr_{(x',y')\sim (\frac{1}{k}\D_j,w^*)} \(\abs{\(y'x'-yx\)\cdot w}\leq \tau \) < \frac{1-\beta}{k}\) < \frac{\beta}{k}.
%\end{equation*}
Summing the probability over all $j\in[k]$, we have that
\begin{equation*}
	\Pr_{(x,y)\sim\D}\( \Pr_{(x',y')\sim\D} \(\abs{\(y'x'-yx\)\cdot w}\leq \tau \) \geq \frac{1-\beta}{k}\) \geq 1-\beta.
\end{equation*}
Note that if this condition holds for all unit vectors, then it also holds for all $w\in\W$. 
%such that $\onenorm{w}\leq\sqrt{s},\twonorm{w}\leq1$.
Hence, $\D$ satisfies $(\tau,\frac{1-\beta}{k},\beta)$-dense pancake condition.
% by taking $\rho=\frac{1-\beta}{k}$.
\end{proof}

The following theorem is an extension of Theorem~18 in~\cite{talwar2020error}. Part of the proof follows from theirs and is included for completeness.
%For the following theorem, we note that part of its proof follows from \cite{talwar2020error}, which we included for completeness.

\begin{theorem}[Restatement of Theorem~\ref{thm:pancake-sample-complexity}]\label{thm:pancake-sample-complexity-restate}
Consider a distribution $\D$ over $\X\times\Y$ satisfying Assumption~\ref{ass:distribution-marginal}  and \ref{ass:dataset-margin} with $\gamma>2\tau$, where $\tau=2\sigma\cdot\(\log \frac{1}{\beta}+1\)$. It holds with probability at least $1-\delta'$ over the draw of $\abs{S_C}\geq \frac{2}{\rho}\cdot\({C_0s}\cdot \log^4{d} +\log\frac{1}{\delta'\beta'}\)$ samples, $(S_C,\D)$ satisfies $\(2\tau,\rho,\beta+\beta'\)$-dense pancake condition, where $C_0>0$ is an absolute constant and $\rho=\frac{1-\beta}{2k}$. 
%Consider a distribution $\D$ over $\X\times\Y$ satisfying Assumption~\ref{ass:dataset-margin} and \ref{ass:distribution-marginal} with $\gamma^*>\tau$, where $\tau=\tau'+\bar{\gamma}$ and  $\tau'=2\sigma\cdot\(\log \frac{1}{\beta}+1\)$. It holds with probability at least $1-\delta'$ over the draw of $n_2\geq \frac{2}{\rho}\cdot\(\frac{C_0s}{\bar{\gamma}^2}\cdot \log\frac{2d}{s} +\log\frac{1}{\delta'\beta'}\)$ samples, $(S,\D)$ satisfies $\(\tau,\rho,\beta+\beta'\)$-dense pancake condition, where $\rho=\frac{1-\beta}{2k}$. 
\end{theorem}
\begin{proof}
Recall that $\rho=\frac{1-\beta}{2k}$.  
From Lemma~\ref{lem:dense-distribution}, we have that $\D$ satisfies $(\tau,2\rho,\beta)$-dense pancake condition. Follow the first part of the proof for Theorem~18 in~\cite{talwar2020error}, we have that 
\begin{equation}\label{eq:density-fix-w}
\Pr_{S\sim \D^n} \(  \Pr_{(x,y)\sim\D}\(  \frac{1}{n}\sum_{i\in S} \one\(\abs{\(y_ix_i-yx\)\cdot w}\leq\tau\) <\rho \) >\beta'+\beta \) \leq \frac{1}{\beta'}\cdot\exp\(-\frac{\rho n}{2}\)
\end{equation}
holds for any fixed $w\in\W:=\{w:\onenorm{w}\leq\sqrt{s},\twonorm{w}\leq1\}$.
We say that a sample $S$ is $(\tau,\rho,\beta)$-good for some $w$ under $\D$ if $\Pr_{(x,y)\sim\D}\(  \frac{1}{n}\sum_{i\in S} \one\(\abs{\(y_ix_i-yx\)\cdot w}\leq\tau\) <\rho \) \leq \beta$. Eq.~\eqref{eq:density-fix-w} implies that a sample $S$ from $\D^n$ is $(\tau,\rho,\beta'+\beta)$-good with respect to a fixed $w\in\W$, except with probability $\frac{1}{\beta'}\cdot\exp(-\frac{\rho n}{2} )$. 

%$S$ is good for all $w\in\W$ with high probability.
Next, we want to show that with a slight relaxation on the pancake thickness, $S$ is not only good for a fixed $w$ but also food for a set of $w'$ close by.
%if $S$ is $(\tau,\rho,\beta'+\beta)$-good with respect to $w$, then $S$ is $(2\tau,\rho,\beta'+\beta)$-good for a $w'$ that is close to $w$. 
Note that for any $(x,y)\in\X\times\Y$, if $i\in S\cap\P_{w}^\tau(x,y) $, we have that $\abs{(y_ix_i-yx)\cdot w} \leq \tau$. 
%\begin{equation*}
%	\abs{(y_ix_i-yx)\cdot w} \leq \tau.
%\end{equation*}
Consider pancake $\P_{w'}^{2\tau}(x,y)$. We know that $S\cap\P_{w}^\tau(x,y) \subseteq S\cap\P_{w'}^{2\tau}(x,y)$ if $\forall i\in S$, %$\abs{(y_ix_i-yx)\cdot(w'-w)}\leq\tau$ 
\begin{equation}\label{eq:pancake-cover}
\abs{(y_ix_i-yx)\cdot(w'-w)}\leq\tau, 
\end{equation}
due to the triangle inequality. It then remains to construct a $\tau$-net of the inner product function space $F:=\{f_w(\bar{x})=w\cdot\bar{x}: \onenorm{w}\leq\sqrt{s},\infnorm{\bar{x}}\leq 2\sigma\cdot\log(d/\delta')\}$ (recall that $\forall i$, $\infnorm{y_ix_i-yx}\leq2\sigma\cdot\log(d/\delta')$ for $(x_i,y_i)$ and $(x,y)$ from the same component, except for probability $\delta'$) and then take a union bound. According to Corollary~5 in~\cite{zhang2002cover} (which we put in Lemma~\ref{lem:cover-number} for reference), we have that the $\infty$-norm covering number of $F$ is upper bounded by $\log \N_{\infty}(F,\tau,n)\leq O\(\frac{\sqrt{s}^2\cdot (\sigma\cdot\log(d/\delta'))^2(2+\log d)}{\tau^2}\cdot\log(\frac{\sqrt{s}\cdot \sigma\cdot\log(d/\delta')}{\tau}\cdot n) \) = O(s\log^4 d)$, due to the setting of $\sigma$ and $\tau$. Note that we need an $\infty$-norm covering number because Eq.~\eqref{eq:pancake-cover} must hold for all $i\in S$ (see more details in~\cite{zhang2002cover}). As a result, we have that 
\begin{align*}
&\Pr_{S\sim \D^n} \( \exists w\in\W:  \Pr_{(x,y)\sim\D}\(  \frac{1}{n}\sum_{i\in S} \one\(\abs{\(y_ix_i-yx\)\cdot w}\leq 2\tau\) <\rho \) >\beta'+\beta \) \\
&\leq \frac{1}{\beta'}\cdot\exp\(C_0\cdot s\log^4 d-\frac{\rho n}{2}\)
\end{align*}
where $C_0>0$ is some constant.
%Then, we aim to show that when the above condition holds for a $w$, it also holds with respect to any $w'$ that is close to $w$ with a slightly larger pancake size. That is, we want to show that the whole pancake $\P^{\tau'}_{{w}}(x,y)$ is fully covered by another pancake $\P^{\tau'+\bar{\gamma}}_{w'}(x,y)$ for any $(x,y)$.  For any $(x',y') \in \P^{\tau'}_{{w}}(x,y)$, it holds that
%\begin{equation*}
%\abs{(y'x'-yx)\cdot w} \leq \tau'.
%\end{equation*}
%Hence, 
%\begin{align*}
%\abs{(y'x'- yx)\cdot w'} &\leq \abs{(y'x'-yx)\cdot(w'-w)} +  \abs{(y'x'-yx)\cdot w} \\
%&\leq \abs{(y'x'-yx)\cdot(w'-w)} + \tau' \\
%&\leq \twonorm{y'x'-yx}\twonorm{w'-w} + \tau'.
%\end{align*}
%It is known that $\twonorm{y'x'-yx}\leq C_5$ from the logconcave concentration and considering that $(x',y')$ is in the pancake. Hence, it suffices to do an $(\frac{\bar{\gamma}}{C_5})$-covering of the hypothesis class $\W$. Due to the result from~\cite{plan2013robust}, we know that the covering number
%\begin{equation*}
%\log N\(\W,\epsilon\)  \leq \frac{C_6}{\epsilon^2}\cdot\width(\W),
%\end{equation*}
%where $\width(\W)$ is the Gaussian mean width of set $\W$ and is bounded by $s\log\(\frac{2d}{s}\)$. Hence,
%\begin{align*}
%\Pr_{S\sim \D^n} \(  \Pr_{(x,y)\sim\D}\(  \frac{1}{n}\sum_{i\in S} \one\(\abs{\left(y_ix_i-yx\right)\cdot w}\leq\tau'+\bar{\gamma}\) <\rho \) >\beta'+\beta \) \\
%\leq \frac{1}{\beta'}\cdot\exp\(-\frac{\rho n}{2}\) \cdot \exp\(\frac{C_0s}{\bar{\gamma}^2}\cdot \log\frac{2d}{s} \).
%\end{align*}
Bounding the above failure probabililty with $\delta'$, it suffices to choose
\begin{equation}
n \geq \frac{2}{\rho}\cdot\({C_0s}\cdot \log^4{d} +\log\frac{1}{\delta'\beta'}\).
%n \geq \frac{2}{\rho}\cdot\(\frac{C_0s}{\bar{\gamma}^2}\cdot \log\frac{2d}{s} +\log\frac{1}{\delta'\beta'}\).
\end{equation}

\end{proof}

%%%%%%%%%L1 covering content - Start
%\begin{lemma}\label{lem:new-width-def}
%Let $W\subseteq\R^d$ be a symmetric set. That is, if $w=(w_1,\dots,w_d)\in W$, then $w^{\{-i\}} := (w_1,\dots,w_{i-1},-w_i,w_{i+1},\dots,w_d) \in W$. For any symmetric $W$ and any $g\in\R^d$, let $\hat{w} \leftarrow \arg\max_{w\in W}\inner{g}{w}$, it holds that $\forall i\in[d]$, $g_i\cdot \hat{w}_i \geq 0$.
%%\begin{equation*}
%%\hat{w} \leftarrow \sup_{w\in W}\inner{g}{w}
%%\end{equation*}
%\end{lemma}
%\begin{proof}
%We show it by contradition. Let us first assume that $\exists i\in[d]$ such that $g_i\cdot \hat{w}_i < 0$. Then, we have
%\begin{align*}
%\inner{g}{\hat{w}} &= g_1\hat{w}_1 + \dots + g_{i-1}\hat{w}_{i-1} + g_i\hat{w}_i + g_{i+1}\hat{w}_{i+1} + \dots + g_d\hat{w}_d \\
%&< g_1\hat{w}_1 + \dots + g_{i-1}\hat{w}_{i-1} - g_i\hat{w}_i + g_{i+1}\hat{w}_{i+1} + \dots + g_d\hat{w}_d \\
%&= \inner{g}{\hat{w}^{\{-i\}}}.
%\end{align*}
%Since $\hat{w}^{\{-i\}} \in W$, $\hat{w}$ could not be optimal. Here is the contradition.  The proof is complete.
%\end{proof}
%%%%%%%%%L1 covering content - End

\begin{lemma}[Restatement of Lemma~\ref{lem:clean-sample-complexity}]\label{lem:clean-sample-complexity-restate}
	If we draw a sample $S$ from $\EX$ with size $\abs{S}\geq\frac{3}{\eta_0}\cdot\log\frac{1}{\delta'}$, then it is guaranteed with probability $1-\delta'$ that $\abs{S_C} \geq (1-2\eta_0)\abs{S}$ and $\abs{S_D} \leq 2\eta_0\abs{S}$.
\end{lemma}
\begin{proof}
	It is known that every time the algorithm request a sample from $\EX$, with probability $\eta\leq\eta_0$, it will be a malicious sample $(x,y)\in S_D$. Hence, by Chernoff bound (Lemma~\ref{lem:chernoff}) we have that
	\begin{equation*}
		\Pr\(\abs{S_D}\geq 2\eta_0\abs{S}\) \leq \exp\(-\frac{\eta_0\abs{S}}{3}\).
	\end{equation*}
	Let the probability bounded by $\delta'$, we conclude that it suffices to choose $\abs{S}\geq\frac{3}{\eta_0}\cdot\log\frac{1}{\delta'}$.
\end{proof}

%The following lemma dedicates to the sample complexity for Algorithm~\ref{alg:soft}.

%Without loss of generality, we prove the  lemma with $\onenorm{H}\leq\kappa^2$ and then replace it with $\kappa={\sqrt{s}}$.
%${\color{red}\nuclearnorm{H}\leq1}$

The following proof follows the structure from that of Proposition~27 in~\cite{shen2021attribute}. However, we note that this proof has been adapted to the mixture of logconcaves with non-zero centers and covariance $\sigma^2I_d$. In fact, Lemma~\ref{lem:orlicz-extension} and \ref{lem:orlicz-extension-distribution} in Section~\ref{sec:orlicz-supp} are strong evidences for that learning under mixture of logconcaves is significantly different from that under a single isotropic logconcave. For example, most of the previous results on isotropic logconcave only work for learning of homogeneous halfspaces.
\begin{lemma}[Restatement of Lemma~\ref{lem:soft-sample-complexity}]\label{lem:soft-sample-complexity-restate}
For any distribution $\DX$ satisfying Assumption~\ref{ass:distribution-marginal}, let $S_C=\{x_1,\dots,x_{\abs{S_C}}\}$ be a set of i.i.d. unlabeled instances drawn from $\DX$. Denote $G(H) := \frac{1}{\abs{S_C}}\sum_{i\in S_C}x_i^{\top}Hx_i - \E_{x\sim\DX}[x^{\top}Hx]$. Then with probability $1-\delta$,
\begin{equation*}
	\sup_{H\in\M}\abs{G(H)} \leq 2\(\frac{1}{d}+r^2\)
	%\sup_{H:\onenorm{H}\leq s}\abs{G(H)} \leq 2\(\frac{1}{d}+r^2\)
\end{equation*}
given that $\abs{S_C}\geq {\Theta\(s^2\cdot \log^5d\log^2\frac{1}{\delta} \)}$. 
%There exists an absolute value $C>0$ such that the following holds for any distribution $\DX$ satisfying Assumption~\ref{ass:distribution-marginal}. Let $S=\{x_1,\dots,x_n\}$ be a set of i.i.d. unlabeled instances drawn from $\DX$. Denote $G(H) := \frac{1}{n}\sum_{i=1}^{n}x_i^{\top}Hx_i - \E_{x\sim\DX}[x^{\top}Hx]$. Then with probability $1-\delta$,
%\begin{equation*}
%\sup_{H:\onenorm{H}\leq s}\abs{G(H)} \leq 2\(\frac{1}{d}+r^2\)
%%C\cdot\(\frac{(\log d+1)^2}{d} + r^2\)
%\end{equation*}
%given that $n\geq {\Theta\(s^2\cdot \log^5(nd)\log^2\frac{1}{\delta} \)}$. 
\end{lemma}
\begin{proof}
Recall that $\M:=\{H:H\succeq0,\onenorm{H}\leq s,\nuclearnorm{H}\leq1\}$. We aim to use Adamczak's bound (Lemma~\ref{lem:adamczak}) for a function class $\F := \{x\mapsto x^{\top}Hx: H\in\M \}$. For the ease of presentation, we further let that $\abs{S_C}=n$ throughout this lemma.

%We will use the Adamczak's bound (Lemma~\ref{lem:adamczak}).
%Define function class $\F := \{x\mapsto x^{\top}Hx: \onenorm{H}\leq s \}$. 

We find a function $F(x)$ that upper bounds $f(x)$ for any $f\in\F$. Given that $\abs{f(x)}=\abs{x^{\top}Hx} = \abs{\sum_{k,l} H_{k,l}x^{(k)}x^{(l)}} \leq \(\sum_{k,l}\abs{H_{k,l}}\) \cdot\infnorm{x}^2 \leq s\infnorm{x}^2$, we define  $F(x)= s\infnorm{x}^2$.

We first bound the Orlicz norm for $\max_{1\leq i\leq n}F(x_i)$. According to Lemma~\ref{lem:orlicz}, we have that
\begin{equation*}
\norm{\max_{1\leq i\leq n}F(x_i)}_{\psi_{0.5}} = \norm{\sqrt{\max_{1\leq i\leq n}F(x_i)}}_{\psi_1}^2.
\end{equation*}
Since $F(x_i)=s\infnorm{x_i}^2$, $\forall 1\leq i\leq n$, we have that ${\max_{1\leq i\leq n}F(x_i)} =  s\cdot \max_{1\leq i\leq n} \infnorm{x_i}^2 $. Moreover, $\sqrt{\max_{1\leq i\leq n}F(x_i)} = \sqrt{s} \cdot \max_{1\leq i\leq n} \infnorm{x_i}$.
Hence, Lemma~\ref{lem:orlicz} and \ref{lem:orlicz-extension} implies that 
\begin{align*}
\norm{\sqrt{\max_{1\leq i\leq n}F(x_i)}}_{\psi_1} &= \norm{\sqrt{s}\cdot \max_{1\leq i\leq n} \infnorm{x_i}}_{\psi_1} \\
&=\sqrt{s} \cdot  \norm{\max_{1\leq i\leq n} \infnorm{x_i}}_{\psi_1} \\
&\leq C_2 \cdot \sqrt{s} \sigma \(\frac{r}{\sigma}+1\) \log(nd).
\end{align*}
Therefore, we conclude that %due to part~\ref{part:norm-transfer} of Lemma~\ref{lem:orlicz},
\begin{equation}\label{eq:orlicz-0.5-bound}
\norm{\max_{1\leq i\leq n}F(x_i)}_{\psi_{0.5}} \leq \(C_2 \cdot \sqrt{s}\sigma \(\frac{r}{\sigma}+1\) \log(nd)\)^2.
\end{equation}

Next, we bound the second term in Adamczak's bound, which includes term $\sup_{f\in\F} \E_{x\sim \DX}[(f(x))^2]$. It is equivalent that taking the supremum over $f\in\F$ and taking it over $H\in\M$.
%$H\in \{H: \onenorm{H}\leq s \}$. 
Given the definition of $F(x)$, it is easy to see that $(f(x))^2\leq (F(x))^2 = s^2\infnorm{x}^4, \forall f\in\F$. Applying Lemma~\ref{lem:orlicz}, we have that
\begin{equation*}
\E_{x\sim \DX}[\infnorm{x}^4] =\norm{\infnorm{x}}_4^{4} \leq \norm{\infnorm{x}}_{\psi_4}^{4} \leq \(4!\norm{\infnorm{x}}_{\psi_1} \)^{4},
\end{equation*}
together with the Orlicz norm bound in Lemma~\ref{lem:orlicz-extension-distribution}, we have that
\begin{align}
\sup_{f\in\F} \E_{x\sim \DX}[(f(x))^2] &\leq \E_{x\sim \DX} \left[ (F(x))^2 \right] \notag\\
&=s^2 \cdot \E_{x\sim \DX} \left[ \infnorm{x}^4 \right] \notag\\
%&\leq \kappa^4 \E_{x\sim D} \left[ 8\infnorm{x-\mu}^4 + 8\infnorm{\mu}^4 \right] \\
%&\leq \kappa^4 \cdot \(\E_{x\sim D} \left[ 8\infnorm{x-\mu}^4 \right] +8r^4 \)
&\leq s^2 \(4!\norm{\infnorm{x}}_{\psi_1}\)^4 \notag\\
&\leq (4!C_2)^4 \cdot s^2 \sigma^4\(\frac{r}{\sigma}+1\)^4\log^4(d). \label{eq:sup-exp}
\end{align}

The final step is to use Rademacher analysis to bound $\E_{x\sim \DX^n}\(\sup_{f\in\F}\abs{\frac{1}{n}\sum_{i=1}^{n}f(x_i)-\E_{x\sim \DX}(f(x))} \)$. Let $a=\{a_1,\dots,a_n\}$ where $a_i$'s are indepedent Rademacher variables. By Lemma~26.2 of~\cite{shwartz2014understand}, we have that
\begin{align*}
%\E_{S\sim D^n}\(\sup_{H}\abs{G(H)} \)
&\E_{x\sim \DX^n}\(\sup_{f\in\F}\abs{\frac{1}{n}\sum_{i=1}^{n}f(x_i)-\E_{x\sim \DX}(f(x))} \) \\
&\leq\frac{2}{n}\E_{S,a} \(\sup_{f\in\F}\abs{\sum_{i=1}^{n}a_if(x_i)}\) \\
&\leq\frac{2}{n} \E_{S,a}\( \sup_{\onenorm{H}\leq s}\abs{\sum_{i=1}^{n}a_i x_i^{\top}Hx_i} \)
\end{align*}
Fix a sample set $S$ and consider the randomness in $a$. Here we will use vectorization for a matrix $H$, denoted as $\vectorize(H)$. We have that
\begin{align*}
&\E_{a}\( \sup_{\onenorm{H}\leq s} \abs{\sum_{i=1}^{n}a_ix_i^{\top}Hx_i} \) \\
&\leq \E_{a}\(\sup_{H:\onenorm{\vectorize(H)}\leq s} \abs{\sum_{i=1}^{n}a_i\inner{\vectorize(H)}{\vectorize(x_ix_i^{\top})}} \) \\
&\leq s\sqrt{2n\log(2d^2)} \cdot \max_{1\leq i\leq n}\infnorm{\vectorize(x_ix_i^{\top})} \\
&\leq s\sqrt{2n\log(2d^2)} \cdot \max_{1\leq i\leq n}\infnorm{x_i}^2,
\end{align*}
where the second inequality is due to Lemma~\ref{lem:rademacher-l1}. We then consider expectation over the choice of $S$
\begin{align*}
&\E_{S,a} \( \sup_{\onenorm{H}\leq s} \abs{\sum_{i=1}^{n}a_ix_i^{\top}Hx_i} \) \\
&\leq s\sqrt{2n\log(2d^2)} \cdot \E_{S}\( \max_{1\leq i\leq n}\infnorm{x_i}^2 \) \\
&\leq s\sqrt{4n\log(2d)} \cdot \(2!\cdot\norm{\max_{1\leq i\leq n}\infnorm{x_i}}_{\psi_1}\)^2 \\
&\leq s\sqrt{4n\log(2d)} \cdot 16\cdot C_2^2\cdot \sigma^2\(\frac{r}{\sigma}+1\)^2\log^2(nd),
\end{align*}
where the second inequality is due to Part~\ref{part:between-norms} of Lemma~\ref{lem:orlicz} and the third inequality is due to Lemma~\ref{lem:orlicz-extension}.

As a result, we have that 
\begin{align*}
&\E_{S,a} \(\sup_{f\in\F}\abs{\sum_{i=1}^{n}a_if(x_i)}\) \\
&\leq C_3\cdot\sqrt{n\log d}\cdot s\sigma^2\(\frac{r}{\sigma}+1\)^2\log^2(nd) 
\end{align*}
for some constant $C_3>0$.
In addition,
\begin{align}
&\E_{x\sim \DX^n}\(\sup_{f\in\F}\abs{\frac{1}{n}\sum_{i=1}^{n}f(x_i)-\E_{x\sim \DX}(f(x))} \) \notag\\
&\leq \frac{C_3\cdot\sqrt{\log d}}{\sqrt{n}} \cdot s\sigma^2\(\frac{r}{\sigma}+1\)^2\log^2(nd). \label{eq:exp-sup}
\end{align}

Combining all three steps, i.e. Eq.~\eqref{eq:exp-sup}, \eqref{eq:sup-exp}, \eqref{eq:orlicz-0.5-bound}
\begin{align*}
\sup_{H\in\M}\abs{G(H)} \leq
%\sup_{\onenorm{H}\leq s}\abs{G(H)} \leq 
C_4\cdot\(s\sigma^2\(\frac{r}{\sigma}+1\)^2 \log^2(nd)\cdot \(\sqrt{\frac{\log d}{n}}+\sqrt{\frac{\log \frac{1}{\delta}}{n}}+\frac{\log^2 \frac{1}{\delta}}{n} \) \).
\end{align*}
Let this quantity be upper bounded by some $t>0$, then we get that it suffices choose $n$ as follows,
\begin{equation*}
n=\Theta\(\frac{1}{t^2}\cdot s^2\sigma^4\(\frac{r}{\sigma}+1\)^4\log^4(nd) \cdot \(\log d+\log^2 \frac{1}{\delta}\) \).
\end{equation*}

Let $t=(\sigma+r)^2$, and since $n=O(d)$, it suffices to choose 
\begin{equation}\label{eq:sample-soft}
n= \Theta\(s^2\cdot \log^5(nd)\log^2\frac{1}{\delta} \) = \Theta\(s^2\cdot \log^5(d)\log^2\frac{1}{\delta} \).
\end{equation}
Recall that $\sigma=\frac{1}{\sqrt{d}}$, and so $t = \(\frac{1}{\sqrt{d}}+r\)^2 \leq 2\(\frac{1}{d}+r^2\)$.

%If let $\sigma=\frac{1}{\sqrt{d}}$, $r={O}(\sigma\cdot\log\frac{d}{\beta})$, and $t = \frac{(\log d+1)^2}{d} + r^2$, we have that
%\begin{align}
%n&=\Theta\( \frac{1}{t^2} \cdot s^2\sigma^4\(\frac{d}{\beta}+1\)^4 \log^5(nd)\log^2\frac{1}{\delta} \) \\
%&= \Theta\(s^2\cdot \log^5(nd)\log^2\frac{1}{\delta} \). \label{eq:sample-soft}
%\end{align}
\end{proof}

%\Jie{It remains to bound $\E[x\top H x]$, which will tell us the value $\bar{\sigma}^2$ in the SDP. There is where you need to be thoughtful on whether $\ell_1$-norm alone  suffices.} \Shiwei{This is updated. 20250425}

%The value of $\bar{\sigma}$ in Algorithm~\ref{alg:soft} is determined by the variance of clean instances projected onto matrix $H$. %Given that $\twonorm{w^*}=1$ and $\zeronorm{w^*} \leq s$, we have that $\onenorm{w^*{w^*}^{\top}} \leq \onenorm{w^*}^2 \leq s$. The variance bound is given by $\sup_{\onenorm{H}\leq s} \sum_{i\in S}q_ix_i^{\top}Hx_i$.

The following lemma shows that for the underlying marginal distribution, which is a uniform mixture of $k$ logconcaves, the variance with respect to matrix $H\in\M$ is upper bounded. This is crucial in setting the value of parameter $\bar{\sigma}$ in Algorithm~\ref{alg:soft}.
\begin{lemma}[Restatement of Lemma~\ref{lem:clean-variance}]\label{lem:distribution-variance-bound}
	%Given distribution $\DX$, it holds that $\sup_{H\succeq0,\onenorm{H}\leq\kappa^2, {\nuclearnorm{H}\leq1} }\E_{X\sim\DX}(x^{\top}Hx) \leq 2\(\frac{1}{d}+r^2\)$.
	Given distribution $\DX$ satisfying Assumption~\ref{ass:distribution-marginal}, it holds that $\sup_{H\in\M }\E_{X\sim\DX}(x^{\top}Hx) \leq 2\(\frac{1}{d}+r^2\)$.
\end{lemma}
\begin{proof} 
	Recall that $\M:=\{H\succeq0,\onenorm{H}\leq s,\nuclearnorm{H}\leq1\}$, which is the constraint set of program in Algorithm~\ref{alg:soft}.
	Consider that $H$ is a positive semidefinite matrix with trace norm at most $1$, we can eigendecomposite $H=\sum_{i=1}^{d}\lambda_iv_iv_i^{\top}$ where $\lambda_i\geq0$ and $\sum_{i=1}^{d}\lambda_i\leq1$. Hence, for any $H\in\M$,
	\begin{equation*}
		x^{\top}Hx 
		= \sum_{i=1}^{d}\lambda_ix^{\top} v_iv_i^{\top}x 
		= \sum_{i=1}^{d}\lambda_i (v_i^{\top}x)^2 
		\leq \sum_{i=1}^{d}\lambda_i\left[2(v_i^{\top}(x-\mu_j))^2 +2(v_i^{\top}\mu_j)^2\right] 
	\end{equation*}
	where $\mu_j$ is the mean of the $j$-th logconcave distribution to which $x$ belongs. Without loss of generatliy, we fix a $j$ and the analysis follows for all $j\in[k]$. Then, for any $H\in\M$,
	\begin{align*}
		\E_{x\sim \D_j}[x^{\top}Hx] &\leq \E_{x\sim \D_j} \sum_{i=1}^{d}\lambda_i\left[2(v_i^{\top}(x-\mu_j))^2 +2(v_i^{\top}\mu_j)^2\right] \\
		&= 2\sum_{i=1}^{d}\lambda_i\left( v_i^{\top}\E_{x\sim \D_j}[(x-\mu_j)(x-\mu_j)^{\top}]v_i +(v_i^{\top}\mu_j)^2\right) \\
		&\leq 2\sum_{i=1}^{d}\lambda_i \(\frac{1}{d}\twonorm{v_i}^2 + \twonorm{\mu_j}^2 \) \\
		&\leq 2\(\frac{1}{d} + r^2 \).
	\end{align*}
	
	By linearity of the expected values, we conclude that $\sup_{H\in\M}\E_{x\sim\DX}(x^\top Hx) \leq 2\(\frac{1}{d}+r^2\)$. The proof is complete.
\end{proof}

\begin{proposition}[Restatement of Proposition~\ref{prop:feasibility}]\label{prop:feasibility-restate}
Given that Asssumption~\ref{ass:distribution-marginal} holds, if we draw a sample set $S$ from $\EX$ with size $\abs{S}\geq \Omega(s^2\cdot \log^5d\log^2\frac{1}{\delta'}+\frac{1}{\eta_0}\cdot\log\frac{1}{\delta'})$, with probability at least $1-2\delta'$, the semidefinite program in Algorithm~\ref{alg:soft} is feasible. 
\end{proposition}
\begin{proof}
	To show that the program is feasible, we want to show that there exists a $q^*$ that satisfies all three constraints. From Lemma~\ref{lem:clean-sample-complexity}, we know that $\abs{S_C} \geq (1-2\eta_0)\abs{S}$ with probability $1-\delta'$ if $\abs{S}\geq\frac{3}{\eta_0}\cdot\log\frac{1}{\delta'}$. From Lemma~\ref{lem:soft-sample-complexity} and \ref{lem:clean-variance}, we know that if $\abs{S_C}\geq\Omega\(s^2\cdot \log^5d\log^2\frac{1}{\delta'}\)$, it holds with probability $1-\delta'$,
	\begin{equation*}
		\sup_{H\in\M} \frac{1}{\abs{S_C}}\sum_{i\in S_C}  x_i^{\top}Hx_i \leq 4\(\frac{1}{d}+r^2\) = \bar{\sigma}^2 .
	\end{equation*}
	Let $\eta_0\leq\frac14$, we have that $\abs{S_C}\geq \frac{1}{2}\abs{S}$. Taking the union bound, the above conditions are satisfied with probability at least $1-2\delta'$ if we choose $\abs{S}\geq \Omega\(s^2\cdot \log^5(d)\log^2\frac{1}{\delta'}+\frac{1}{\eta_0}\cdot\log\frac{1}{\delta'}\)$.
	Hence, by setting $q^*_i=1$ for every $i\in S_C$ and setting all other weights to $0$, $q^*$ satisfies all three constraints. 
	%We have showed that the program is feasible.
\end{proof}

\section{Orlicz Norm and Concentration Results}\label{sec:orlicz-supp}

\begin{definition}[Orlicz norm]
	For any $z\in\R$, let $\psi_\alpha: z \mapsto \exp(z^\alpha)-1$. Furthermore, for a random variable $Z\in\R$ and $\alpha>0$, define $\norm{Z}_{\psi_\alpha}$, the Orlicz norm of $Z$ with respect to $\psi_\alpha$, as:
	\begin{equation*}
		\norm{Z}_{\psi_\alpha} = \inf\{t>0:\E_{Z}[\psi_\alpha(\abs{Z}/t)] \leq1 \}.
	\end{equation*}
\end{definition}

The following follows from Section~1.3 of~\cite{van1996weak}.
\begin{lemma}\label{lem:orlicz}
Let $Z, Z_1, Z_2$ be real-valued random variables. Consider the Orlicz norm with respect to $\phi_\alpha$. We have the following:
\begin{enumerate}
	\item $\norm{\cdot}_{\psi_\alpha}$ is a norm. For any $\alpha\in\R$, $\norm{\alpha Z}_{\psi_\alpha} = \abs{\alpha}\cdot\norm{Z}_{\psi_\alpha}$; $\norm{Z_1 + Z_2}_{\psi_\alpha} \leq \norm{Z_1}_{\psi_\alpha} + \norm{Z_2}_{\psi_\alpha}$.
	\item $\norm{Z}_p \leq \norm{Z}_{\psi_p} \leq p!\norm{Z}_{\psi_1}$ where $\norm{Z}_{p} := \(\E[\abs{Z}^p]\)^{1/p}$. \label{part:between-norms}
	\item For any $p,\alpha >0$, $\norm{Z}_{\psi_p}^\alpha = \norm{Z^\alpha}_{\psi_{p/\alpha}}$. \label{part:norm-transfer}
	\item If $\Pr(\abs{Z} \geq t) \leq K_1\exp(-K_2 t^\alpha)$ for any $t\geq0$, then $\norm{Z}_{\psi_\alpha} \leq \(\frac{2(\log K_1+1)}{K_2}\)^{1/\alpha}$. \label{part:orlicz-bound}
	\item If $\norm{Z}_{\psi_\alpha} \leq K$, then for all $t\geq0, \Pr(\abs{Z}\geq t) \leq 2\exp\(-\(\frac{t}{K}\)^{\alpha}\)$.
\end{enumerate}
\end{lemma}

\begin{lemma}\label{lem:orlicz-extension}
There exist absolute constants $C_1,C_2>0$ such that the following holds for any  distribution $\DX$ satisfying Assumption~\ref{ass:distribution-marginal}.
%with mean $\twonorm{\mu}\leq r$ and covariance matrix $\Sigma \prec \sigma^2 I_d$. 
Let $S=\{x_1,\dots,x_n\}$ be a set of $n$ samples from $\DX$. We have that,
\begin{align*}
%&\quad\norm{\max_{x\in S}\infnorm{x-\mu}}_{\psi_1} \leq C_1\cdot \sigma\log(nd);\\
&\norm{\max_{x\in S}\infnorm{x}}_{\psi_1} \leq C_2\cdot \sigma\(\frac{r}{\sigma}+1\)\log(nd).
\end{align*}
In addition,
\begin{equation*}
\E_{S\sim \DX^n}\(\max_{x\in S}\infnorm{x}\) \leq C_1\cdot \sigma\log(nd) + r.
\end{equation*}
\end{lemma}
\begin{proof}
Let $Z$ be zero-mean $\sigma^2$-variance logconcave random variable in $\R$. Lemma~\ref{lem:logconcave} implies that
\begin{equation}\label{eq:logconcave-tail}
\Pr\(\abs{Z} > t\) \leq \exp(-t/\sigma +1).
\end{equation}
Since $\DX$ is a mixture of $k$ logconcave distribution, we denote by $\bar{x}_i$ the $i$-th instance subtracting its logconcave distribution mean, e.g. $(x_i-\mu_j)$ if $x_i$ is from $\D_j,j\in[k]$.
Fix an instance $i\in\{1,\dots,n\}$ and a coordinate $l\in\{1,\dots,d\}$. Denote by $\bar{x}_i^{(l)}$ the $l$-th coordinate of $\bar{x}_i$. Then, we have that Eq.~\eqref{eq:logconcave-tail} holds for any $\bar{x}_i^{(l)}$. Taking the union bound over all instances and all coordinates gives us that,
\begin{equation*}
\Pr_{S\sim D^n}\(\max_{x\in S}\infnorm{\bar{x}} > t \) \leq nd\cdot\exp\(-\frac{t}{\sigma}+1\).
\end{equation*}
Applying  Part~\ref{part:orlicz-bound} of Lemma~\ref{lem:orlicz}, we have that
\begin{equation*}
\norm{\max_{x\in S}\infnorm{\bar{x}}}_{\psi_1} \leq C_1\cdot\sigma\log(nd).
\end{equation*}
Likewise, for each coordinate $\bar{x}_i^{(l)}$, we have that $\Pr\(\abs{\bar{x}_i^{(l)}} > t\) \leq \exp(-t/\sigma +1)$, which implies that  $\Pr\(\abs{x_i^{(l)}} > t\) \leq \exp(-(t-r)/\sigma +1)$ for all $t>r$, since $\abs{\mu_j^{(l)}} \leq \twonorm{\mu_j}\leq r, \forall j\in[k]$. For $S\sim D^n$, we have that $\Pr(\max_{x\in S}\infnorm{x} > t) \leq nd\cdot\exp(r/\sigma+1)\exp(-t/\sigma)$.
We use similar technique to bound the Orlicz norm for $\max_{x\in S}\infnorm{x}$ as follows,
\begin{equation*}
\norm{\max_{x\in S}\infnorm{x}}_{\psi_1} \leq \frac{2\(\(r/\sigma+1\)\log(nd)+1\)}{1/\sigma} = C_2\cdot \sigma\(\frac{r}{\sigma}+1\)\log (nd).
\end{equation*}

On the other hand, we have that
\begin{equation*}
\E_{S\sim D^n}\(\max_{x\in S}\infnorm{x}\) \leq \E_{S\sim D^n}\(\max_{x\in S}\infnorm{\bar{x}} + r\) = \E_{S\sim D^n}\(\max_{x\in S}\infnorm{\bar{x}}\)  + r.
%\E_{S\sim D^n}\(\max_{x\in S}\infnorm{x}\) \leq \E_{S\sim D^n}\(\max_{x\in S}\infnorm{x-\mu} + \infnorm{\mu}\) = E_{S\sim D^n}\(\max_{x\in S}\infnorm{x-\mu}\)  + \infnorm{\mu}.
\end{equation*}
%In addition, we have that 
since $\infnorm{\mu} \leq \twonorm{\mu} \leq r$. 
%Hence, $\E_{S\sim D^n}\(\max_{x\in S}\infnorm{x}\) \leq E_{S\sim D^n}\(\max_{x\in S}\infnorm{x-\mu}\)+r$.

The lemma then follows from the fact that $\E_{S\sim D^n}\(\max_{x\in S}\infnorm{\bar{x}}\) \leq \norm{\max_{x\in S}\infnorm{\bar{x}}}_{\psi_1}$.
\end{proof}

\begin{lemma}\label{lem:orlicz-extension-distribution}
There exist absolute constants $C_1,C_2>0$ such that the following holds for any logconcave distribution $D$ satisfying Assumption~\ref{ass:distribution-marginal}.
%There exists an absolute constant $C_1,C_2>0$ such that the following holds for any logconcave distribution $D$ with mean $\twonorm{\mu}\leq r$ and covariance matrix $\Sigma \prec \sigma^2 I_d$,
\begin{align*}
%	&\quad\norm{\infnorm{x-\mu}}_{\psi_1} \leq C_1\cdot \sigma\log(d);\\
	&\norm{\infnorm{x}}_{\psi_1} \leq C_2\cdot \sigma\(\frac{r}{\sigma}+1\)\log(d).
\end{align*}
In addition,
\begin{equation*}
	\E_{x\sim D}\big(\infnorm{x}\big) \leq C_1\cdot \sigma\log(d) + r.
\end{equation*}
\end{lemma}
\begin{proof}
The lemma follows from the analysis in Lemma~\ref{lem:orlicz-extension} with  $n=1$ sample.
\end{proof}

%\Shiwei{until here.}

\section{Lower Bounds}\label{sec:lower-bound-app}

In this section, we show the sample complexity lower bounds for attribute-efficient PAC learning of sparse halfspaces. We use some arguments in prior works of~\cite{shen2021attribute} and \cite{dia2017statistical} and show that our sample complexities are indeed nearly optimal.
Specifically, Lemma~2 of~\cite{shen2021attribute} provided an information-theoretic lower bound of label complexity for attribute-efficiently active learning of $s$-sparse halfspaces under single isotropic logconcave marginals.
We use similar arguments to show a sample complexity lower bound of passive learning under the mixture of logconcaves.

\cite{benedek1988learn} proved that a concept class $W$ is $(\epsilon,\delta)$-learnable with respect to a fixed distribution $\D_\X$ with at least $\log ((1-\delta)N(2\epsilon,W,\D_\X))$ samples, where $N(\epsilon,W,\D_\X)$ is the $\epsilon$-covering number of $W$ under the metric of $(W,\D_\X)$. In more details, this metric measures the distance of two concepts $w_1,w_2\in W$ using $\Pr_{(x,y)\sim\D_\X}(w_1(x)\neq w_2(x))$. In~\cite{kulkarni1993active}, it was shown that the lower bound for the label complexity of active learning is quantitatively the same.
%$\dis_{\D_\X}(c_1,c_2)=$
\cite{shen2021attribute} then utilizes this result together with known lower bounds of $\epsilon$-packing numbers of $s$-sparse halfspaces, i.e. $(\frac{1}{\epsilon})^{\Omega(s)}$~\cite{long1995on} and $(\frac{d}{s})^{\Omega(s)}$~\cite{raskutti2011minimax}, to derive the lower bound under isotropic logconcaves. Notice that the result implicitly depends on a fact that the Euclidean distance or the angle between two unit vectors $u,v$ implies the error rate distance under isotropic logconcave distribution, which we include as follows.
%Lemma~10 of~\cite{zhang2018efficient} and
%Lemma~3 of~\cite{balcan2013active} 
\begin{lemma}[Lemma~3 of~\cite{balcan2013active}]\label{lem:angle-error}%[Euclidean implies error rate distance]
	For any two unit vectors $u,v\in\R^d$, there are absolute constants $c > 0$ such that for any isotropic logconcave distributions $D$, $c\cdot\theta(u,v) \leq \Pr_{x\sim D} \( \sign (u \cdot x) \neq \sign (v \cdot x) \)$.
	%For any two vectors $u,v\in\R^d$, there are absolute constants $c_1, c_2 > 0$ such that the following holds for all isotropic logconcave distributions $D$, 
	%\begin{equation}
	%c_1 \cdot \Pr_{x\sim D} \(\sign (u \cdot x) = \sign (v \cdot x)\) \leq \theta(u,v) \leq c_2 \cdot \Pr_{x\sim D} \( \sign (u \cdot x) = \sign (v \cdot x) \)
	%\end{equation}
\end{lemma}
Inspired by Lemma~10 of~\cite{zhang2018efficient}, we also summarize the following lemma.
\begin{lemma}\label{lem:angle-euclidean}
	For any two unit vectors $u,v$, we have that $\frac{2}{\pi}\cdot \theta(u,v) \leq \twonorm{u - v} \leq \theta(u,v)$.
\end{lemma}
\begin{proof}
	By Jordan's inequality, we have that $\frac{2}{\pi}\cdot \theta \leq \sin\theta \leq \theta$ for any $\theta\in [0,\frac{\pi}{2}]$. Together with the fact that $\frac{\twonorm{u-v}}{2}=\sin\frac{\theta(u,v)}{2}$, we obtain the bounds.
\end{proof}
Combining Lemma~\ref{lem:angle-error} and \ref{lem:angle-euclidean}, the result in \cite{shen2021attribute} is obtained. We now show that this bound also holds for mixtures of logconcaves. This requires adapting the proof of Lemma~\ref{lem:angle-error} from isotropic logconcave to mixture of logconcaves. In this paper, we only show the difference and refer interested readers to the proof of Lemma~3 of~\cite{balcan2013active}. To that end, we have the following theorem.
\begin{lemma}[Restatement of Lemma~\ref{lem:lower-bound}]\label{lem:lower-bound-restate}
	Learning of $s$-sparse halfspaces under Assumption~\ref{ass:dataset-margin} for any $\gamma\leq\frac12$ and \ref{ass:distribution-marginal} in the noiseless setting has an information-theoretic sample complexity lower bound of $\Omega(s\cdot(\log\frac{1}{\epsilon}+\log\frac{d}{s}))$.
	%Learning of $s$-sparse halfspaces under Assumption~\ref{ass:dataset-margin} and \ref{ass:distribution-marginal} in the realizable case has an information-theoretic sample complexity lower bound of $\Omega(s\cdot(\log\frac{1}{\epsilon}+\log\frac{d}{s}))$.
\end{lemma}
\begin{proof}
	It suffices to construct a hardness result that there exist a mixture of logconcave distributions $D$ under which $(\epsilon,\delta)$-learning of $w^*$ requires at least $\Omega(s\cdot(\log\frac{1}{\epsilon}+\frac{d}{s}))$ samples. Notice that the $\epsilon$-packing number is at least $(\frac{1}{\epsilon})^{\Omega(s)}+(\frac{d}{s})^{\Omega(s)}$. It remains to show that under distribution $D$, $\exists c>0$ such that $c\cdot \twonorm{u-v} \leq \Pr_{x\sim D} \( \sign (u \cdot x) \neq \sign (v \cdot x) \)$. 
	
	Define  $\proj_{u,v}(D)$ to be the distribution given by first picking $x\sim D$ and then orthogonally projecting $x$ onto the plane determined by $u$ and $v$. \cite{balcan2013active} showed that $\Pr_{x\sim D} \( \sign (u \cdot x) \neq \sign (v \cdot x) \) = \Pr_{x\sim \proj_{u,v}(D)} \( \sign (u \cdot x) \neq \sign (v \cdot x) \)$. Moreover, let $A$ be the region of disagreement between $u$ and $v$ intersected with the ball of radius $1/9$ in $\R^2$ defined by $u$ and $v$. Then, $\Pr_{x\sim \proj_{u,v}(D)} \( \sign (u \cdot x) \neq \sign (v \cdot x) \) \geq \vol(A)\cdot \inf_{x\in A}\proj_{u,v}(D)(x) \geq c\cdot\theta(u,v)$, where $\vol(A)$ is the volume of $A$ and $\proj_{u,v}(D)(x)$ is the probability density of $x$ under $\proj_{u,v}(D)$.
	%Since each component of $D$ is a logconcave, which preserves the logconcavitivity after projection. Hence, $\proj_{u,v}(D)$ is also a mixture of logconcaves. Without loss of gener
	
	Now, consider $D$ to be a distribution which has $0$ probability density inside the region of $B:=\{x:\abs{w^*\cdot x}\geq\gamma\}$ and uniform outside it, i.e. $\bar B$. Note that this $D$ satisfies both Assumption~\ref{ass:dataset-margin} and \ref{ass:distribution-marginal}. We will show that there is still at least $(\frac{1}{\epsilon})^{\Omega(s)}+(\frac{d}{s})^{\Omega(s)}$ number of $s$-sparse halfspaces that are $\epsilon$ away from each other, i.e. $\Pr_{x\sim D} \( \sign (u \cdot x) \neq \sign (v \cdot x) \)\geq \epsilon$. To show this, consider any unit vectors $u,v$ such that $\theta(u,w^*)\geq\frac{\pi}{4}$ and $\theta(v,w^*)\geq\frac{\pi}{4}$. Now,  $\Pr_{x\sim \proj_{u,v}(D)} \( \sign (u \cdot x) \neq \sign (v \cdot x) \) \geq \vol(A\backslash B)\cdot \inf_{x\in A\backslash B}\proj_{u,v}(D)(x) \geq c\cdot\theta(u,v)\cdot (1-(\sqrt{2}\gamma)^2)\geq\frac{c}{2}\theta(u,v)$ for $\gamma\leq\frac12$. Here, the second transition is due to that $A\cap B$ is contained in a sector with angle $\theta(u,v)$ and radius $\leq\sqrt{2}\gamma$. 
	%The third transition is because $\gamma\leq\frac12$ for large enough $d$.
	%it suffices to show that $\vol(\bar B)\geq c\cdot\vol(B\cup \bar B)$ for some constant $c>0$. To see this, consider two
	
	In other words, for any unit vector $u$ such that $\theta(u,w^*)\geq\frac{\pi}{4}$, the Euclidean distance of $c\cdot \epsilon$ implies $\Pr_{x\sim D} \( \sign (u \cdot x) \neq \sign (v \cdot x) \) \geq \epsilon$. Define $A':=\{u\in\R^d:\zeronorm{u}\leq s,\theta(u,w^*)\geq\frac{\pi}{4}\}$. It is not hard to notice that $\vol(A')$ is propotional to that of $W$. Hence, the packing number stays the same. By taking $\log N(2\epsilon,A',D) = \Omega(s\cdot(\log\frac{1}{\epsilon}+\log\frac{d}{s}))$ we  obtain the sample complexity lower bound as presented in the lemma. The proof is complete.
\end{proof}

Last but not least, it was well known that for malicious noise model (or Huber contamination model in unsupervised learning), there exists a statistical-computational gap of $\Omega(s^2)$ for any statistical-query algorithm~\cite{dia2017statistical}. This is a strong evidence showing that our sample complexity is almost optimal.
% between the information-theoretic sample lower bound and that

\section{Attribute-Efficient Learning with Adversarial Label Noise}\label{sec:adv}

The algorithm for attribute-efficient learning under a constant adversarial label noise is as follows.

\begin{algorithm}
	\caption{Main Algorithm}\label{alg:adversarial}
	\begin{algorithmic}[1]
		\REQUIRE $\EX(D, w^*, \eta)$, target error rate $\epsilon$, failure probability $\delta$, parameters $s,\gamma$.
		\ENSURE A halfspace $\hat{w}$.
		\STATE Draw $n'={ C\cdot {s^2}\log^5\frac{d}{\delta\epsilon} }$ samples from $\EX_\text{adv}(D, w^*, \eta)$ to form a set ${S'}=\{(x_i,y_i)\}_{i=1}^{n'}$.
		\STATE Remove all samples $(x,y)$ in $S'$ with $\infnorm{x}\geq r+\sigma\cdot(\log\frac{n'd}{\delta'}+1)$ to form a set $S$.
		\STATE Solve the following hinge loss minimization program:
		\begin{equation}
			\hat{w} \leftarrow \underset{\onenorm{w}\leq {\sqrt{s}}, \twonorm{w}\leq1}{\arg\min} \ell_\gamma(w; S).
		\end{equation}
		\STATE Return $\hat{w}$. 
	\end{algorithmic}
\end{algorithm}

We will use the natural concentration of instances from the mixture of logconcave distributions and show that a bounded variance allows Algorithm~\ref{alg:adversarial} to return a good enough halfspaces $\hat{w}$ that has a low error rate with high probability.

\begin{definition}[Learning sparse halfspaces with adversarial label noise]\label{def:adv-noise}
Given any $\epsilon,\delta\in(0,1),$ and an adversary oracle $\EX_\text{adv}(\D, w^*, \eta)$ with underlying distribution $\D$, ground truth $w^*$ with $\zeronorm{w^*}\leq s$, and noise rate $\eta$ fixed before the learning task begins, when the learner requests a set of samples from $\D$, the oracle draws a set $S_0\sim\D^n$ and then flips an arbitrary $\eta$ fraction of the labels in $S_0$, and returns the modified set $S$ to the learner.
%everytime the learner requests a sample from $\EX(\D, w^*, \eta)$, with probability $1-\eta$, the oracle returns an $(x,y)$ where $(x,y)\sim\D$ and $y=\sign(x\cdot w^*)$; with probability $\eta$, the oracle returns an arbitrary sample $(x,y)\in\X\times\Y$. The goal of the learner is to output a halfspace $\hat{w}$ using a number $n=\poly(s,\log d)$ of samples such that the error rate $\err_{\D}(\hat{w})\leq\epsilon$ with probability $1-\delta$, where $\err_{\D}(w):=\Pr_{(x,y)\sim\D}(y\neq\sign(w\cdot x))$.
%The marginal distribution of $\D$ on $\X$ satisfies Assumption~\ref{ass:distribution-marginal}, and it satisfies that $\Pr_{(x,y)\sim\D}(\sign(w^*\cdot x)\neq y)\leq \eta$.
\end{definition}

\begin{theorem}[Attribute-efficent learning under adversarial label noise]\label{thm:main-adv}
Given a set $S$ of $n\geq\Omega \({s}\cdot\log^4\frac{d}{\delta\epsilon} \) $ samples from $\EX_\text{adv}(\D, w^*, \eta)$ with $\D,w^*$ satisfying Assumption~\ref{ass:dataset-margin}  with $\gamma\geq {\frac{4(\log\frac{1}{\epsilon} +1)}{\sqrt{d}} }$ and Assumption~\ref{ass:distribution-marginal} with $r\leq 2\gamma$, $k\leq 64$, and the adversarial label noise rate $\eta\leq\eta_0\leq \frac{1}{2^{32}}$, the following holds. For any $\epsilon,\delta\in(0,1)$, Algorithm~\ref{alg:adversarial} returns a $\hat{w}$  such that $\err_{\D}(\hat{w})\leq\epsilon$ with probability at least $1-\delta$.
\end{theorem}
\begin{proof}[Proof of Theorem~\ref{thm:main-adv}]
We again consider that $S$ consists of two sets of samples. i.e. $S_C \cup S_D$ where $S_C$ are the samples not being corrupted, and $S_D$ are those whose labels are flipped. 
We also consider the set $\bar{S}$ of instances without labels from set $S$.

We show that similar deterministic guarantees to Theorem~\ref{thm:deterministic-restate} holds for Algorithm~\ref{alg:adversarial}. 
%That is, we aim to show that if Assumption~\ref{ass:dataset-margin} holds,  Assumption~\ref{ass:pancake-deterministic} holds with $\tau\leq\frac{\gamma}{2}$, and it further holds that
%\begin{equation}\label{eq:new-rho}
%\rho\geq {32\(\frac{1}{\gamma\sqrt{d}}+\frac{r}{\gamma}+1\)\sqrt{\eta_0}},
%\end{equation}
%then, the output $\hat{w}$ of Algorithm~\ref{alg:adversarial} has error rate $\err_{\D}(\hat{w})\leq \epsilon$.
\begin{claim}
If Assumption~\ref{ass:dataset-margin} holds, and
%Assumption~\ref{ass:pancake-deterministic} holds with $\tau\leq\frac{\gamma}{2}$
\begin{equation}\label{eq:new-rho}
	c\cdot\rho\geq {\(\frac{1}{\gamma\sqrt{d}}+\frac{r}{\gamma}+1\)\sqrt{\eta_0}}
\end{equation}
for some some constant $c\in(0,\frac{1}{32})$, then, the output $\hat{w}$ of Algorithm~\ref{alg:adversarial} has error rate $\err_{\D}(\hat{w})\leq \epsilon$.
\end{claim}
Note that $\D$ satisfies some dense pancake condition (Lemma~\ref{lem:dense-distribution}), and $S_C$ is obtained by removing at most $\eta$ fraction from $S$.
%, and the instance set $\bar{S}$ is an uncorrupted instance set from $\DX$. 
Hence, due to Theorem~\ref{thm:pancake-sample-complexity-restate}, it holds with probability at least $1-\delta'$ that $(S_C,\D)$ satisfies the $(\tau,\rho-\eta,\epsilon)$-dense pancake condition with respect to any $w\in\W$ with $\abs{S} \geq\Omega\( \frac{1}{\rho}\cdot\({s \cdot \log^4 d}+\log\frac{1}{\delta'\epsilon}\) \)$,
%\begin{equation*}
%\abs{S} \geq\Omega\( \frac{1}{\rho}\cdot\(\frac{s \cdot \log d}{\bar{\gamma}^2}+\log\frac{1}{\delta'\epsilon}\) \)
%\end{equation*}
where $\tau=4\sigma\cdot\(\log \frac{1}{\epsilon}+1\)$, $\rho=\frac{1-\epsilon}{2k}$. It is easy to see from Eq.~\eqref{eq:new-rho} that $\eta\leq\eta_0\leq\frac{\rho}{2}$. Thus, $(S_C,\D)$ satisfies $(\tau,\rho/2,\epsilon)$-dense pancake condition. We choose that $\gamma\geq 2\tau =  \frac{8(\log\frac{1}{\epsilon} +1)}{\sqrt{d}}$. 
In the following, we rescale $\rho'=\rho/2$.
%Within small rescaling factor of $\rho$, the analysis for Theorem~\ref{thm:deterministic-restate} holds under Eq.~\eqref{eq:new-rho}.

%similar to Theorem~\ref{thm:local-correctness}, 
Then, we make the following claim.
%, which is a counterpart of Theorem~\ref{thm:local-correctness} for the adversarial noise.

\begin{claim}\label{cl:local-correctness-adv}
Due to Assumption~\ref{ass:dataset-margin}, for any given $(x,y)$, if its pancake $\P_{\hat{w}}^{\tau}(x,y)$ with $\tau\leq {\gamma}/{2}$ is $\rho'$-dense with respect to $S_C$ for some $\rho'>4\eta_0$, and it holds that %$\frac14\cdot\sum_{i\in S_C\cap \P_{\hat{w}}^{\tau}(x,y)} q_i > \frac{\sqrt{s}}{\gamma}\cdot\linsum(q\circ S_D)$, 
\begin{equation}\label{eq:density-note-rate-adv}
	\abs{S_C\cap \P_{\hat{w}}^{\tau}(x,y)} > \frac{4}{\gamma}\cdot\linsum(S_D),
\end{equation}
then $(x,y)$ is not misclassified by the $\hat{w}$ returned by Algorithm~\ref{alg:adversarial}. 
\end{claim}
%We first bound $\linsum(S_D)$ using an existing technique in~\cite{talwar2020error}. That is, observe that by definition, the gradient norm in Definition~\ref{def:linsum} is smaller than linear sum norm  in their work. By their Theorem~21, we see that %$\linsum(S_D)\leq C\cdot\(\sqrt{\abs{S_D}}+\abs{S_D}\cdot r + \frac{\abs{S_D}\log\frac{\abs{S}}{\abs{S_D}}}{\sqrt{d}}\)$ 
%\begin{align}
%\linsum(S_D) &\leq C\cdot\(\sqrt{\abs{S_D}}+\abs{S_D}\cdot r + \frac{\abs{S_D}\log\frac{\abs{S}}{\abs{S_D}}}{\sqrt{d}}\) \\
%&\leq C\cdot\(\sqrt{\eta\abs{S}} + \eta\abs{S}\cdot \(2\gamma + \frac{\log1/\eta}{\sqrt{d}} \) \)
%\end{align}
%for some large enough constant $C>0$.
The proof follows similarly from that of Theorem~\ref{thm:local-correctness}. Notice that we do not require reweighting the samples here, hence, we can reproduce the proof by setting $q$ to an all-one vector. Moreover, we consider the gradient norm on $S_D$ directly. 
We then show that Eq.~\eqref{eq:density-note-rate-adv} holds under the condition of Eq.~\eqref{eq:new-rho}
, which is similar to proving Lemma~\ref{lem:weight-clean}. 
We note that $\abs{S_C\cap \P_{\hat{w}}^{\tau}(x,y)} \geq (\rho-\eta)\abs{S}$, since $\P_{\hat{w}}^{\tau}(x,y)$ is assumed to be $\rho$-dense and $\abs{S_D}$ takes up to $\eta$-fraction (Definition~\ref{def:adv-noise}). 

Then, we show that $\linsum(S_D)\leq 2\sqrt{\frac{1}{d}+r^2} \cdot \sqrt{\eta}\abs{S}$. That is, we apply Lemma~\ref{lem:linsum-bound-restate} that $\linsum(S_D) \leq \sqrt{\abs{S_D}}\sqrt{\sup_{w\in\W}\sum_{i\in S_D}(w\cdot x_i)^2}$ by letting the weight vector $q$ be all ones. In addition, $\sup_{w\in\W}\sum_{i\in S_D}(w\cdot x_i)^2 \leq \sup_{w\in\W}\sum_{i\in S} (w\cdot x_i)^2 \leq 4(\frac{1}{d}+r^2)\cdot\abs{S}$, where the second transition is due to Lemma~\ref{lem:soft-sample-complexity-restate} and \ref{lem:distribution-variance-bound} with the fact that $ww^{\top}\in\M$. Note that this is satisfied with probability $1-\delta'$ by choosing $\abs{S}\geq\Theta\(s^2\cdot \log^5d\log^2\frac{1}{\delta'} \)$.The gradient norm is well bounded. Finally, it is easy to obtain Eq.~\eqref{eq:density-note-rate-adv} from Eq.~\eqref{eq:new-rho}.

%Finally, it suffices to let 
%\begin{equation*}
%\rho - \eta > \frac{4C}{\gamma}\cdot \(\sqrt{\frac{\eta}{\abs{S}}} +\eta \cdot \(2\gamma + \frac{\log1/\eta}{\sqrt{d}} \) \),
%\end{equation*}
%that is,
%\begin{equation*}
%\rho > 4C \cdot \( \frac{\sqrt{\eta}}{\gamma\sqrt{\abs{S}}} +3\eta \)
%\end{equation*}

Therefore, given  Claim~\ref{cl:local-correctness-adv} holds, Eq.~\eqref{eq:new-rho} holds with $k\leq 64$ and $\eta_0\leq\frac{1}{2^{32}}$, and $(S_C,\D)$ satisfies $(\tau,\rho/2,\epsilon)$-dense pancake condition, we conclude that all underlying instances but an $\epsilon$ fraction is not misclassified by the returned $\hat{w}$. That is, $\err_{\D}(\hat{w})\leq\epsilon$.

It then remains to conclude the sample complexity for determinimistic result and the confidence bound. Let $\delta'=\delta/2$. Then, $\err_{\D}(\hat{w})\leq\epsilon$ holds with probability $1-\delta$. The sample complexity is as presented in the theorem, which is similar to that in Theorem~\ref{thm:main}.

%Then, we want to utilize Theorem~\ref{thm:local-correctness}, which still holds true for the label noise condition. It remains to show Eq.~\eqref{eq:density-noise-rate} holds uner this setting (reproducing Lemma~\ref{lem:weight-clean}). It suffices to bound $\bar{\sigma}$

%$D'$ satisfies $(\tau',2\rho-\eta,\beta+\eta)$-dense pancake due to Lemma~\ref{lem:dense-distribution} and Lemma~16 of~\cite{talwar2020error}. Together with Theorem~\ref{thm:pancake-sample-complexity-restate}, we conclude that $S$ satisfies $(\tau,\rho-\frac{\eta}{2},2\epsilon+\eta)$-dense pancake condition if 
%\begin{equation*}
%\abs{S}\geq\Omega\(   \frac{2}{\rho-\frac{\eta}{2}}\cdot\(\frac{C_0s}{\bar{\gamma}^2}\cdot \log\frac{2d}{s} +\log\frac{1}{\delta'(2\epsilon+\eta)}\)  \)
%\end{equation*}
%where $\tau = \tau'+\bar{\gamma}$, $\tau'=\sigma\cdot\(\log 1/\epsilon+1\)$
%%since the attribute vectors are not corrupted, we have 

\end{proof}

\section{Useful Lemmas}\label{sec:useful_lemmas}

\begin{lemma}[Isotropic logconcave, \cite{LV07geometry}]\label{lem:logconcave}
For any isotropic logconcave distribution $D$ over $\R^d, d\geq1$, it holds that
\begin{equation*}
\Pr_{x\sim D}(\twonorm{x}\geq t\sqrt{d})\leq \exp(-t+1).
\end{equation*}
In addition, any orthogonal projection of $D$ onto subspace of $\R^d$ is still isotropic logconcave.
\end{lemma}

\begin{lemma}[Chernoff bound]\label{lem:chernoff}
Let $Z_1, Z_2, \dots, Z_n$ be $n$ independent random variables that take value in $\{0, 1\}$. Let $Z = \sum_{i=1}^{n} Z_i$. For each $Z_i$, suppose that $\Pr(Z_i =1) \leq \eta$.  Then for any $\alpha \in [0, 1]$
\begin{equation*}
	\Pr\left( Z \geq  (1+\alpha) \eta n\right) \leq e^{-\frac{\alpha^2 \eta n}{3} }.
\end{equation*}
When $\Pr(Z_i =1) \geq \eta$, for any $\alpha \in [0, 1]$
\begin{equation*}
	\Pr\left( Z \leq  (1-\alpha) \eta n\right) \leq e^{-\frac{\alpha^2 \eta n}{2} }.
\end{equation*}
The above two probability inequalities hold when $\eta$ equals exactly $\Pr(Z_i = 1)$.
\end{lemma}

\begin{lemma}[Adamczak's bound]\label{lem:adamczak}
For any $\alpha\in(0,1]$, there exists a constant $\Lambda_\alpha>0$, such that the following holds. Given any function class $\F$, and a function $F$ such that for any $f\in\F$, $\abs{f(x)}\leq F(x)$, we have with probability at least $1-\delta$ over the draw of a set $S=\{x_1,\dots,x_n\}$ of i.i.d. instances from a distribution $D$,
%\begin{equation*}
%\begin{split}
%\sup_{f\in\F} \abs{\frac{1}{n}\sum_{i=1}^{n}f(x_i)-\E_{x\sim D}[f(x)] } 
%\leq 
%\Lambda_\alpha \( \E_{S\sim D^n}\( \sup_{f\in\F} \abs{\frac{1}{n}\sum_{i=1}^{n}f(x_i)-\E_{x\sim D}[f(x)]} \) 
%\\+ \sqrt{\frac{\sup_{f\in\F}\E_{x\sim D}[f^2(x)]\log\frac{1}{\delta}}{n} } 
%+ \frac{(\log\frac{1}{\delta})^{1/\alpha}}{n}\cdot\norm{\max_{1\leq i\leq n}F(x_i)}_{\psi_\alpha}  \),
%\end{split}
%\end{equation*}
\begin{multline*}
\sup_{f\in\F} \abs{\frac{1}{n}\sum_{i=1}^{n}f(x_i)-\E_{x\sim D}[f(x)] } 
\leq 
\Lambda_\alpha \Bigg( \E_{S\sim D^n}\Bigg( \sup_{f\in\F} \abs{\frac{1}{n}\sum_{i=1}^{n}f(x_i)-\E_{x\sim D}[f(x)]} \Bigg) \\
+ \sqrt{\frac{\sup_{f\in\F}\E_{x\sim D}[f^2(x)]\log\frac{1}{\delta}}{n} } 
+ \frac{(\log\frac{1}{\delta})^{1/\alpha}}{n}\cdot\norm{\max_{1\leq i\leq n}F(x_i)}_{\psi_\alpha}  \Bigg),
\end{multline*}
where $\norm{\cdot}_{\psi_\alpha}$ denotes the Orlicz norm with parameter $\alpha$.
\end{lemma}

\begin{lemma}[Theorem~1 of~\cite{kakade2008on}]\label{lem:rademacher-l1}
Let $a=(a_1,\dots,a_n)$ where $a_i$'s are independent draws from the Rademacher distribution and let $x_1,\dots,x_n$ be given instances in $\R^d$. Then,
\begin{equation*}
\E_{a}\(\sup_{\onenorm{w}\leq t}\sum_{i=1}^{n}a_iw\cdot x_i\) \leq t\sqrt{2n\log(2d)}\max_{1\leq i\leq n}\infnorm{x_i}.
\end{equation*}
\end{lemma}

\begin{definition}[$p$-norm covering number]\label{def:covering-number}
Let $F:=\{f_w:\R^d\rightarrow\R, w\in\R^d \}$ be a set of real-valued functions. Consider a set of observations $X^n=\{x_1,\dots,x_n\} \subseteq \R^d$ and a vector $f_w(X^n) = (f_w(x_1), \dots,f_w(x_n)) \in \R^n$ parameterized by $w$, the covering number in $p$-norm, denoted as $\N_p(F,\epsilon,X^n)$, is the minimum number $m$ of a collection of vectors $v_1,\dots,v_m \in\R^n$ such that $\forall w$, there exists $v_j$ satisfying
\begin{equation*}
\norm{f_w(X)-v_j}_p = \( \sum_{i=1}^{n}(f_w(x_i)-v_j^i)^p \)^{1/p} \leq n^{1/p}\cdot \epsilon,
\end{equation*}
where $v_j^i$ denotes the $i$-th component of vector $v_j$. We also define $\N_p(F,\epsilon,n) = \sup_{X^n}\N_p(F,\epsilon,X^n)$.
\end{definition}

\begin{lemma}[Corollary~5 of~\cite{zhang2002cover}]\label{lem:cover-number}
Consider $x,w\in \R^d$. Let $F$ be the set of linear functions on $x$ induced by $w$, i.e. $F:=\{f_w: x\mapsto w\cdot {x}\}$. Let $\N_\infty(F,\epsilon,n)$ be the $\infty$-norm covering number of $F$ as defined in Definition~\ref{def:covering-number}.
%with precision $\epsilon$ and restricting to any set of $x$'s with cardinality no more than $n$. 
If $\infnorm{x}\leq b$ and $\onenorm{w}\leq a$, then $\forall \epsilon>0$,
\begin{equation*}
\log \N_\infty(F,\epsilon,n) \leq \frac{288a^2b^2(2+\ln d)}{\epsilon^2}\log_2\(2\left\lceil \frac{8ab}{\epsilon}+2 \right\rceil n +1\).
\end{equation*}
\end{lemma}

\end{document}